# Analysis of COVID-19 Cases in India through Machine Learning: A Study of Intervention

Hanuman Verma [a], Akshansh Gupta [b] and Utkarsh Niranjan [c]

[a]*Department of Mathematics, Bareilly College, Bareilly, Uttar Pradesh-243005, India*
[b]*CSIR-Central Electronics Engineering Research Institute, Pilani, Rajasthan-333031, India*
[c]*School of Computer and Systems Sciences, Jawaharlal Nehru University, New Delhi-110067, India*
*Email: [hv4231@gmail.com](hv4231@gmail.com)[a], [akshanshgupta@ceeri.res.in](akshanshgupta@ceeri.res.in)[b], [utkarshkumarniranjan@gmail.com](utkarshkumarniranjan@gmail.com)[c]*

**Abstract:** To combat the coronavirus disease 2019 (COVID-19) pandemic, the world has vaccination, plasma therapy, herd immunity, and epidemiological interventions as few possible options. The COVID-19 vaccine development is underway and it may take a significant amount of time to develop the vaccine and after development, it will take time to vaccinate the entire population, and plasma therapy has some limitations. Herd immunity can be a plausible option to fight COVID-19 for small countries. But for a country with huge population like India, herd immunity is not a plausible option, because to acquire herd immunity approximately 67% of the population has to be recovered from COVID-19 infection, which will put an extra burden on medical system of the country and will result in a huge loss of human life. Thus epidemiological interventions (complete lockdown, partial lockdown, quarantine, isolation, social distancing, etc.) are some suitable strategies in India to slow down the COVID-19 spread until the vaccine development. Government of India (GOI) and different state governments have made many efforts to mitigate the COVID-19 spread under various scenarios through the intervention and public awareness. In this work, we have suggested the SIR model with intervention, which incorporates the epidemiological interventions in the classical SIR model. To model the effect of the interventions, we have introduced $\rho$ as the intervention parameter. $\rho$ is a cumulative quantity which covers all type of intervention. We have also discussed the supervised machine learning approach to estimate the transmission rate ($\beta$) for the SIR model with intervention from the prevalence of COVID-19 data in India and some states of India. To validate our model, we present a comparison between the actual and model-predicted number of COVID-19 cases. In most cases, we find the predicted numbers of cases are very much close to the actual number of cases, but there is some mismatch between the predicted and actual number of cases due to irregular imposed intervention, uneven demographic distribution, personal behavior, and individual negligence. Using our model, we also present predicted numbers of active and recovered COVID-19 cases till Sept 30, 2020, for the entire India and some states of India and also estimate the 95% and 99% confidence interval for the predicted cases.

## 1. Introduction

A disease that quickly spreads within a short period to a large population in a certain region is called an epidemic. Epidemics occur because of several internal and external factors such as a genetic change in the pathogen reservoir, a change in the ecology, or the entering the developing pathogen in a host population. An epidemic may be restricted to one particular place or spread to other places in a short period. An epidemic is termed as pandemic if it spreads to several countries and affects a large number of people globally [1]. Today, the world has become small with the availability of faster transportation. People can move from one place to another place in a short time.

In December 2019, coronavirus disease 2019 (COVID-19) was identified in Wuhan, the capital city of China's Hubei province [2]. COVID-19 is characterized as an infectious disease and it occurs due to Severe Acute Respiratory Syndrome Coronavirus 2 (SARS-CoV-2) [3] and it spreads among people through close contact with infected individuals generally via small droplets produced by sneezing, coughing, or talking. COVID-19 can also spread through the infected surface if someone touches the infected surface and touches his/her nose, mouth, or eyes. World Health Organization (WHO) declared the COVID-19 outbreak as a Public Health Emergency of International Concern (PHEIC) on January 30, 2020, and on March 11, 2020, declared as a pandemic of infectious disease [4]. On July 07, 2020, the WHO [5] acknowledged that the evidence emerging of the airborne spread of the COVID-19 in the air under certain conditions.

In India, the first case of COVID-19, who had returned from Wuhan, China, was reported positive on January 30, 2020, in Kerala [6]. Following that two more cases were reported in Kerala on Feb 02, 2020, and Feb 03, 2020. Approximately a month later on March 02, 2020, new cases of COVID-19 were reported. Gradually the disease spread in India and became an epidemic. To fight with the COVID-19 emergency, India has Vaccination, Herd Immunity, Plasma Therapy, and Epidemiological Interventions as few possible options.

Although the COVID-19 vaccine development is underway and it may take a significant amount of time to develop the vaccine [7]. Currently, the University of Oxford, Jhonson & Jhonson, National Institute of Virology (India), Cansino Biologics Inc. (China) are some organizations working to develop the vaccine. Even after the development of the vaccine, the production and dissemination of vaccines is a very challenging task. In India, with an approximate current population of 135 crores, if we vaccinate 10 lakh people per day it will take more than three years to vaccinate the whole population of India. Producing ten lakh vaccines per day and then vaccinating the same number of people is a mammoth task.

Herd immunity or community immunity is an idea from epidemiology. Herd immunity gained more attention for COVID-19 after a remark from England's Prime Minister Boris Jhonson. In herd immunity, a large fraction of the population has acquired immunity against the virus. The fraction of the immune population depends on the value of the reproduction ratio [8]. In various researches [9, 10, 11] it has been found that for COVID-19 an approximately 67% of the population is required to be immune against the disease to gain herd immunity. There are two possible ways to gain herd immunity either after recovery from infection or through vaccination. Presently in the absence of the vaccine, we have the only option of immunity after recovery. Herd immunity can be a plausible option to fight COVID for small countries. But for India, Herd immunity is not a very promising alternative. With a 2-3% fatality rate and herd immunity requirement of 67%, India will face huge human life loss during the course of acquiring herd immunity.

Convalescent plasma therapy (CPT) is also being considered as a treatment of COVID-19. In the CPT, blood plasma from a recently recovered patient is transfused in COVID-19 patients. COVID-19 neutralizing antibodies are present in the donor's plasma increases the recipient's immune response against the disease. As suggested by ICMR this therapy is being used as last resort on the compassionate ground by hospitals in severe COVID-19 cases. The Indian government and state governments have started collecting plasma for the treatment. Finding a suitable donor is a challenging task. Moreover, CPT is being used for severe patients so it is not a very viable treatment option.

So from the above discussion, vaccination is the most effective way to win over COVID-19. Till date there is no vaccine for COVID-19, prevention is the only cure we have. To control the spread of COVID-19, some recommended preventive measures are hand washing, maintaining physical social distance, wearing a face mask in public places, and personal protective equipment (PPE) [12]. The monitoring and self-isolation/quarantined are also recommended for people who feel some symptoms with COVID-19. The worldwide countries have responded by implementing travel restrictions, lockdowns, and school/college and facility closures.

In the absence of the COVID-19 vaccine, the preventive measures, effective treatments, and interventions such as lockdown are a key part of handling the COVID-19, which will help to reduce the new COVID-19 infected cases and delay the peak that is the turning point of COVID-19 cases. This is known as flattening the curve. The interventions reduce the new infected COVID-19 cases which help the healthcare services to better manage the same volume of the patients [13].

India has declared COVID-19 outbreak as an epidemic in some states and union territories, and as per the preventive measures, many commercial institutions and the educational institutions have been shut down. A 14-hour voluntary public curfew at the instance of the honorable prime minister of India has declared on March 22, 2020. Further to control the COVID-19 outbreak, on March 24, 2020, the honorable prime minister announced a first complete lockdown nationwide for 21 days up to April 14, 2020 [14]. On April 14, 2020, the honorable prime minister extended the ongoing nationwide complete lockdown till May 03, 2020. Further, on May 01, 2020, the government of India extended nationwide complete lockdown till May 17, 2020. All the districts of states divided into three zones: green, red, and orange zones based on present and past active COVID-19 cases and some relaxations applied according to the zones [15]. On May 17, 2020, National Disaster Management Authority (NDMA), India further extended lockdown till May 31 [16]. After then the lockdown restrictions were started to lift with some conditions, which is termed as unlocking. The first phase of unlocking, "unlock 1.0" was announced from June 1, 2020, to June 30, 2020, and further "unlock 2.0" declared from July 01, 2020, to July 31, 2020, with more ease of restrictions [17]. In the unlock phase, the services are resumed in a phased manner that is the partial lockdown.

These nationwide lockdown and partial lockdown have slowed the transmission rate of COVID-19 and so delay the peak of the infectious disease, resulting in the less number of COVID-19 cases in India. Complete lockdown, partial lockdowns, or other types of intervention restrictions were as aggressive for containing the spread and building necessary healthcare infrastructure in India. The government of India got time to building special COVID-19 hospital and emergency investment in health care, fiscal stimulus, investment in the vaccine, and drug. But, due to nationwide lockdowns, the Indian economy has been severely affected. The maximum impact of this intervention is expected to be on daily wage workers and those below the poverty line.

In the context of India, India is the second most populous country in the world. As per the India census 2011 data, 68.84 % of India's population lives in rural areas while 31.16 % stay in urban areas. In northeast India, the population density is low in comparison to other states of India. Most of the population residing in rural areas is more isolated and maintained the physical social distance in the comparison of urban areas. The villages in the rural areas are separated by a few kilometers and the rural area markets are not as crowded as compared to urban areas. This helps to maintain the physical social distance in the rural area. During the lockdown period, due to the unavailability of work, migrant laborers started to return to their native places. The maximum of these migrant workers belongs to rural areas. To reduce the risk of spread COVID-19 in rural areas governments decided to impose mandatory 14 days quarantine at a public facility that helped to restrict the disease.

The epidemiological interventions are key measures to curb the COVID-19 spread and reduce new COVID-19 cases. COVID-19 transmission occurs through close contact with infected individuals. A susceptible person can come near a COVID-19 infected person at the workplace, school/college, household, market, crowded place, or in the community. The chance of getting infection depends on the spatial distance between the contacts. Individual personal behavior (social distancing, frequent hand sanitization, and wearing a mask, etc.) also plays a key role to control the COVID-19 spread. Personal rather than government preventive actions are more important to save individuals form infection. In the case of COVID-19 infection, self-isolation, maintaining physical social distancing, and seeking medical advice are very helpful as the symptom appears.

Epidemiological models are very helpful to understand the spread dynamics of an epidemic. These models for infectious disease are used to predict the spread rate of the disease, the duration of the disease, and the peak of the infectious disease. These models can be used for short term and long term predictions for the infectious disease that may be used in decision making to optimize possible controls from the infectious disease. This in turn helps to public health intervention. In literature, several mathematical models for infectious diseases have been introduced. These models categorized into collective models and network models [18] as two main groups. Logistic models [19], generalized growth models [20], Richards's models [21], sub epidemics wave models [22], Susceptible-Infected-Recovered (SIR) model [23], and Susceptible-Exposed-Infectious-Removed (SEIR) models belong in the group of collective models. The SIR model is a compartmental model where the whole population considered a closed population and this closed population is divided into three compartments that are susceptible, infected, and recovered. Some infected persons with diseases are introduced to the population. These infected persons infect some other persons who are susceptible during their sickness at an average rate $R_0$, which is known as the basic reproduction number.

Recently, some works have been reported in the literature using the SIR model to predict the COVID-19 outbreak. Among these works, Sourish Das [24] has estimated the basic reproduction number $R_0$ for the SIR model using statistical machine learning approaches and make the prediction of COVID-19 for India. Ndiaye et al. [25] have analyzed the COVID-19 by the SIR model and machine learning techniques for forecasting the spread of COVID-19 in the world. Further, a comparative prediction of COVID-19 confirmed cases with deterministic and stochastic SIR models and machine learning approaches has been studied by Ndiaye et al. [26]. Liu et al. [27] have studied and predicted the COVID-19 cumulative number of cases in China where the early reported cases are used. Chen et al. [28] introduced the time-dependent SIR model for COVID-19 with the assumption that the transmission rate $(\beta)$ and recovery rate $(\gamma)$ varying with time and estimated these parameters by ridge regression. To determine the COVID-19 transmission rate, a time-dependent state-space SIR model has been proposed by Deo et al. [29] proposed which incorporate the intervention for prediction of COVID-19 in India. In literature, some other works have been reported to incorporate the impact of interventions [7, 30, 31].

In this work, we introduce the impact of the intervention in the SIR model and suggested a model named, the SIR model with intervention to study the COVID-19 spread in India. And, also discussed a machine learning approach to estimate the transmission rate $(\beta)$ for the SIR model with the intervention $(\rho)$ from the prevalence reported COVID-19 cases in India. We determined the intervention value $(\rho)$ with fitting the model predicted COVID-19 cases with actual reported COVID-19 cases. The impact of different levels of interventions is demonstrated in our work. We also present a prediction of COVID-19 active and recovered cases for entire India and some states of India.

The rest of the paper is organized as follows. Section 2, describes the basic SIR model. The proposed SIR model with intervention is presented in section 3. The linear regression method is explained in section 4 and further, the estimation of the transmission rate is defined in section 5. In section 6, we introduce the COVID-19 dataset and demography of India. In section 7, we present our analysis and prediction of the COVID-19 progression in India. Finally, the conclusion has made in section 8.

2.  **SIR Model**

The SIR (Susceptible-Infectious-Recovered) is a compartmental model in epidemiology, introduced by Kermack and McKendrick [23]. It is a mathematical model which tells how an infectious disease spreads through a population during a period. In this model, the whole population size ($N$) of country or location is considered as constant and divided into three compartments with labels Susceptible ($S$), Infectious ($I$), and Recovered ($R$) that varies with the time ($t$). The natural death and birth that is the demography are not considered in the SIR model, because the period of the infectious disease is much shorter than the human lifetime. The people who belong in the population may progress between different compartments, shown in Fig. 1 and corresponding changes occur in terms of two parameters $\beta$ and $\gamma$.

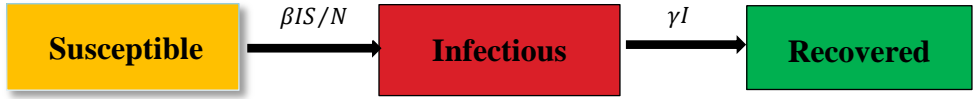

**Fig. 1**: SIR model diagram

The SIR model is described in the following set of differential equations as [23]:

$$\begin{cases} \frac{dS(t)}{dt} = -\frac{\beta\, S(t)\, I(t)}{N} \\ \frac{dI(t)}{dt} = \frac{\beta\, S(t)\, I(t)}{N} - \gamma\, I(t) \\ \frac{dR(t)}{dt} = \gamma\, I(t) \end{cases} \quad (1)$$

where:
- $S(t)$: Number of individuals susceptible but not yet to infected with an infectious disease at time $t$
- $I(t)$: Number of infected individuals at time $t$ and can spread the infectious disease to susceptible individuals
- $R(t)$: Number of recovered (or deceased) individuals at time $t$ and assumed to be immune for life
- $\beta$: Transmission rate through the exposure of the infectious disease
- $\gamma$: Rate of recovering from the infectious disease, and $1/\gamma$ is the mean period during which an infected individual can pass it on

From the Eq. (1), it can be seen that $\frac{dS}{dt} + \frac{dI}{dt} + \frac{dR}{dt} = 0$ which implies that $S + I + R = constant = N$. Each susceptible individual contacts $\beta$ people per day with a fraction $I/N$ which are infectious, and $\beta SI/N$ move out of the susceptible compartment and goes into the infectious compartment. The parameter $\gamma$ is the rate of recovery and $\gamma I$ move out of the infectious compartment and goes into the recovered compartment. $R_0 = \beta/\gamma$ is the basic reproduction number and it denotes the expected number of new infectious from a single infected person in a population where all subjects are susceptible. If $R_0 = 1$, one infected person infects an average of one person that is the spread of infectious disease is stable, and $R_0 = 2$ means a single infected person infects an average of two people. If $R_0 < 1$ this indicates one infected person infects on average less than one person and spread of infectious disease is expected to stop, and if $R_0 > 1$ this means an infected person infects on average more than one person spread of disease is increasing in the absence of an intervention. It is demonstrated in Fig. 2.

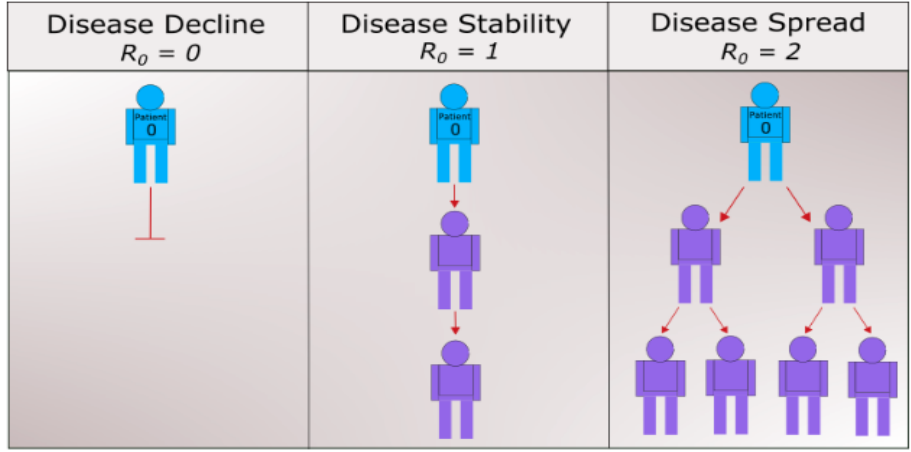

**Fig. 2**: Demonstration of $R_0$ and infectious disease spread [32]

The SIR model tries to predict the total number of infected persons or duration for infectious diseases that are spread from human to human. It is generally run with ordinary differential equations, which is a deterministic model and it can be also used with a stochastic (random) context.

In the SIR model, the determination of the basic reproduction number $R_0 = \beta/\gamma$ is a challenging task for COVID-19 infectious disease. The recovery rate for the COVID-19 infectious disease is considered as on average 14 days [7, 24]. In our study, we have considered the value of $1/\gamma = 14$. To determine the value of $R_0$, we have estimated the transmission rate $(\beta)$, which is described in section 5.

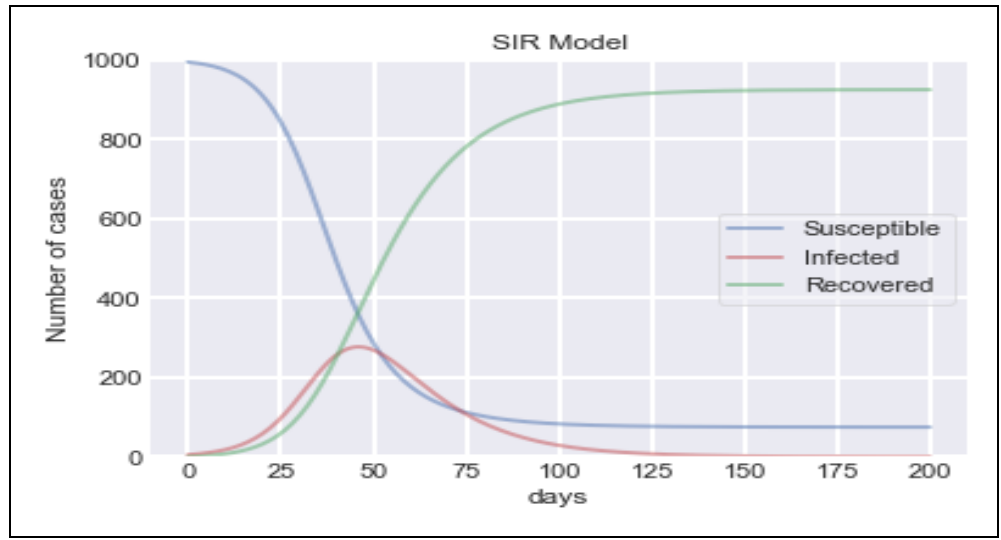

**Fig. 3**: SIR model

Here, we demonstrate the SIR model for a sample population size 1000 with initially five infected persons with the disease. At time $t = 0, N = 1000, I(0) = 5, R(0) = 0$ and $S(0) = N - I(0) - N(0)$. Let the transmission rate $\beta = 0.20$ and mean recovery rate $\gamma = 1/14$ (1/days), which implies the basic reproduction number $R_0 = 2.8$. The SIR model graph with infected, recovered, and susceptible cases is shown in Fig. 3.

Initially, susceptible cases (Fig. 3) decrease slowly that may be the incubation period of infectious disease. It decreases rapidly after 14 days, which shows that the susceptible persons move out from the susceptible compartment and goes into the infectious compartment. The infected persons are shown by the red curve (Fig. 3). Initially, five people exposed in the closed population and with $R_0 = 2.8$, and it infected more than 275 persons per day in 48 days, which is the peak for the infectious disease. The infected persons start recovered from the disease, shown in the green curve (Fig. 3) that is moving out from the infectious compartment to the recovered compartment. After around 42 days the number of recovered people increases as compared to infected people.

3. **SIR Model with intervention**

As we discussed the classical SIR model in section 2, which is based on the assumption that the whole population is closed and does not consider any intervention such as complete lockdown, partial lockdown, isolation, quarantine, and social distancing, etc. In the scenario of the traditional SIR model where no interventions are assumed, infected individuals can infect other people with the transmission rate $\beta$ and COVID-19 infection disease will survive for a longer period and eventually die out as most of the population either have recovered or removed. But in reality, this intervention affects the transmission rate $\beta$ for COVID-19 spread. Suppose in the closed population, if 5 ($I(0) = 5$) person being infected with the COVID-19 and immediately these 5 infected persons tested positive and have been isolated themselves/quarantined from the entire population, then no other person will get the infection from these 5 persons and infection disease may die out. To incorporate this intervention in the SIR model, we introduced the factor $(1 - \rho)$ with transmission rate $\beta$ where $\rho$ is the value of the intervention. $\rho$ is a cumulative quantity which covers all type of intervention. The SIR model with the intervention factor can be defined as:

$$\begin{cases} \frac{dS(t)}{dt} = -\frac{(1-\rho)\,\beta\,S(t)\,I(t)}{N} \\ \frac{dI(t)}{dt} = \frac{(1-\rho)\,\beta\,S(t)\,I(t)}{N} - \gamma\,I(t) \\ \frac{dR(t)}{dt} = \gamma\,I(t) \end{cases} \quad (2)$$

The value of intervention $\rho$ lie in the interval [0,1], where $\rho = 0$ denotes there is no intervention and $\rho = 1$ denotes there is a complete intervention. Other levels of intervention $\rho \in (0, 1)$. For $\rho = 0$, the SIR model with intervention reproduces the classical SIR model. For the case, $\rho = 1$ means there is a complete intervention, the factor $(1 - \rho)$ becomes zero, this implies that $S(t) = constant$ and no further new infected cases will occur. The infected person will be recovering and infection will die soon.

The SIR model with intervention for a sample population size 1000 with initially five COVID-19 infected persons is demonstrated in Fig. 4. Let $\beta = 0.20$, $\gamma = 1/14$ and $t = 0, N = 1000, I(0) = 5, R(0) = 0$ and $S(0) = N - I(0) - N(0)$. The SIR model with different intervention graph is depicted in Fig. 4(a)-(f).

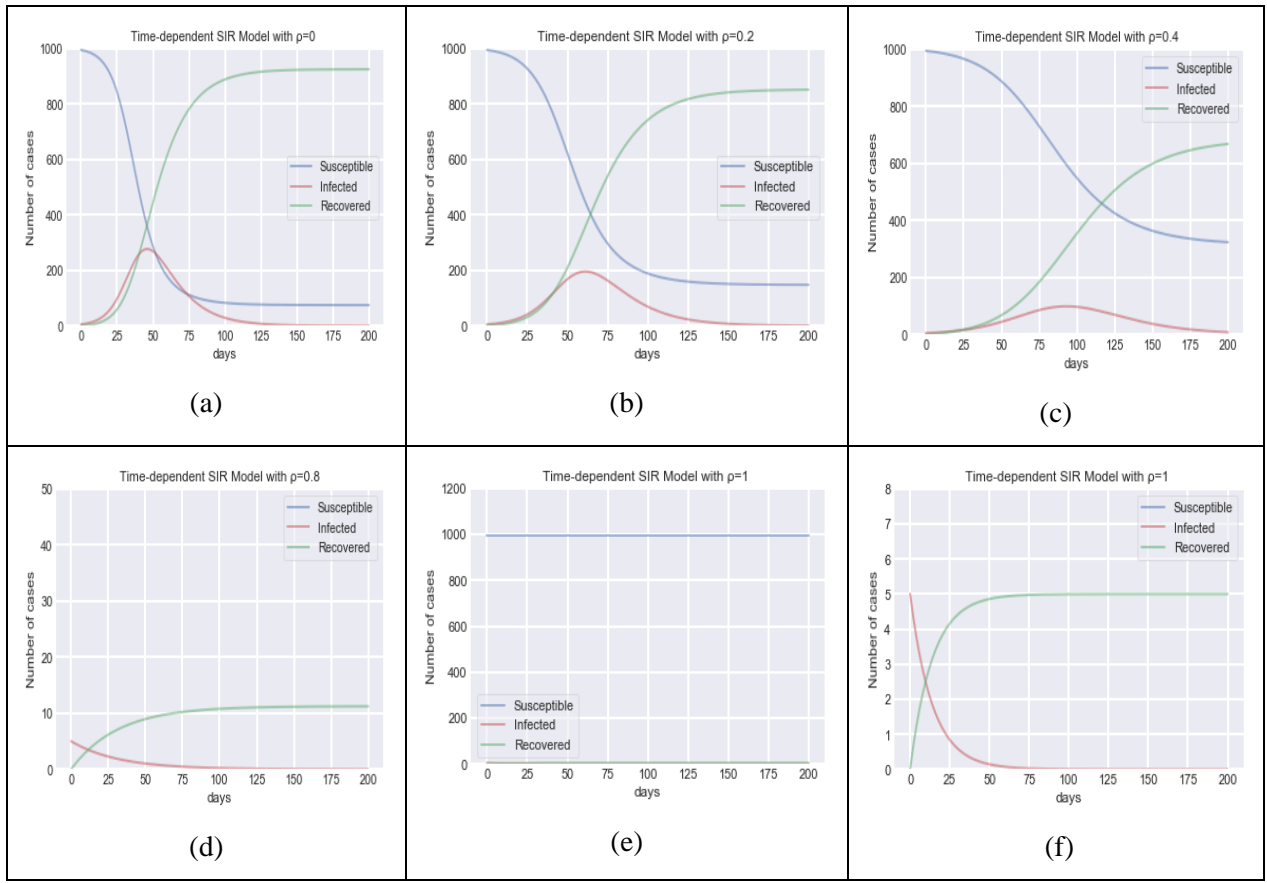

**Fig. 4**: SIR model with intervention for different value of interventions: (a) $\rho = 0$ (b) $\rho = 0.2$ (c) $\rho = 0.4$ (d) $\rho = 0.5$ (e) $\rho = 0.8$ (f) $\rho = 1$

When $\rho = 0$, the SIR model with intervention reduces to the classical SIR model as discussed in section 2, with the same parameters, and also shown in Fig. 4(a). In the case of complete intervention $\rho (= 1)$, the factor $(1 - \rho) = 0$, and from Eq. (2) we obtained $S(t) = constant$, which is also experimentally demonstrated in Fig. 4(e) that is the susceptible cases become constant which is 995. No progress from the susceptible compartment to the infectious compartment and infected person will recover and move out to the recovered compartment, which is shown in Fig. 4(f), where five infected persons recovered in around 25 days and no new infected persons reported. Fig. 4(b) and Fig. 4(c) represents the SIR model with intervention $\rho = 0.2$ and $\rho = 0.4$. For $\rho = 0.2$, the maximum number of newly infected cases is 195 on a single day, which is peak value that is the turning point, which occurred at $62th$ days and disease will die out in 190 days. However, for $\rho = 0.4$, the maximum number of infected cases is 99 on a single day, which is peak value for this case, which occurred at $95th$ days and disease will die out in 262 days. As we take the large intervention value, suppose $\rho = 0.8$, shown in Fig. 4(d), the infectious disease will die out in $50th$ days and the maximum number of infected per day is 5 which the initial infected value at the time $t = 0$. In this case with the considered parameters, it can be observed that as the intervention value $p \in (0, 0.6]$, the number of infected cases per day decreases and peak value also take more days and disease die out takes time. However, when intervention value $p \in (0.6, 1)$, the infectious diseases die out soon and newly infected cases per day also decrease. The small level of intervention slowed the growth of the COVID-19 pandemic and flatting the curve and more level of an intervention force to restrict and pandemic soon die out. The objective of epidemiologists to helping the government's policymakers to minimize the new COVID-19 cases and associated mortality, delay the pandemic peak that will help to manage the health care and effort of the economy at the manageable level to wait for vaccine and manufacturing at large scale. For this how many levels of intervention should manageable.

4. **Linear regression**

Linear regression is a linear approach to find the relationship between one dependent variable or response variable and one independent variable or predictor variable. And, the multiple linear regressions are an approach, which find the linear relationship between one dependent variable and more than one independent variable [33]. Linear regression is widely used in many fields to describe possible relationships between variables. It also used in the in the field of machine learning as one of the supervised machine learning algorithms. The relationships between the variables are defined using linear predictor functions whose unknown model parameters are estimated from the training data. For an independent variable $x$ and a dependent variable $y$ whose relationship is a straight line, the linear regression model is defined as [33];

$$y = \alpha_0 + \alpha_1 x + \varepsilon \qquad (3)$$

where $\alpha_0$ is known as intercept, and $\alpha_1$ is known as slope, are two unknown regression coefficients in the linear regression. $\epsilon$ is a random error with mean zero and variance $\sigma^2$. The estimation of $\alpha_0$ and $\alpha_1$ should be in a straight line, which are the best fit to the considering data. For $n$ pairs of observation ($x_i$, $y_i$), $i = 1,2,3 \dots n$, the linear regression model is defined as [33];

$$y_i = \alpha_0 + \alpha_1 x_i + \varepsilon_i \qquad i = 1,2,3,\dots,n \qquad (4)$$

To make the linear regression algorithm more accurate, the minimization of the sum of least square error between the predicted and actual value is described as:

$$L = \sum_{i=1}^{n} \varepsilon_i^2 = \sum_{i=1}^{n}(y_i - \alpha_0 - \alpha_1 x_i)^2 \qquad i = 1,2,3,\dots,n \qquad (5)$$

5. **Estimation of transmission rate ($\beta$) using linear regression from prevalence COVID-19 data**

For the COVID-19 projection, transmission rate ($\beta$) is a difficult task to determine directly which perfectly fit for the SIR model with intervention. In literature, some works have been reported to measures the parameters from the early data with the inverse problem in epidemiology [34, 35, 36].

To estimate transmission rate ($\beta$), we have considered prevalence reported COVID-19 infected cases from India. In the deterministic case, the differential equation $\frac{dI(t)}{dt}$ is written as $\frac{dI(t)}{dt} \approx \frac{\Delta I(t)}{\Delta t}$, where $\Delta I(t) = I(t+1) - I(t)$ and $\Delta t = (t+1) - t$ days. Here $I(t)$ is the cumulative sum of infected active cases during the time of course and $\Delta I(t)$ are the infected active persons per day. From the differential Eq. (1), it is written as:

$$\frac{dI(t)}{dt} \approx \frac{\Delta I(t)}{\Delta t} = \frac{\beta\, S(t)\, I(t)}{N} - \gamma\, I(t) \qquad (6)$$

After simplifying, it can be written as:

$$\beta = \left(\frac{\Delta I(t)}{\Delta t} \times \frac{1}{I(t)} + \gamma\right)\frac{N}{S(t)} \qquad (7)$$

With prevalence COVID-19 data, we have determined the linear equation, where time/days are independent variable and $\beta$ are the dependent variables, using linear regression to estimate the value of transmission rate ($\beta$) for the COVID-19 prediction. The sklearn, package of Python is used for estimating $\beta$ value using a linear model and 10 cross-validations are utilized to obtain optimal value of $\beta$.

## 6. COVID-19 data and demographics of India

### 6.1 COVID-19 Data

In this study, we obtained the COVID-19 data from COVID-19India.org [37] reported between June 05, 2020, and July 25, 2020. The number of confirmed, recovered, and deceased cases for states wise and India are publicly available at COVID-19India [37]. It is a crowd-sourced open database for COVID-19 available online: https://api.covid19india.org/documentation/csv/

### 6.2 Demographics of India

India is the second most populous country in the world and it occupies 2.41% of the world land area and accommodates 17.7 % World's population. As per the India census 2011, India's population is 121 crore. Of 121 crore Indians, 68.84 % of India's population lives in rural areas while 31.16 % stays in urban areas, which are tabulated in Table 1 and also depicted in Fig. 5. India's population density as per the census 2011 is 382 per square kilometer. Bihar is the most densely populated state (1102 persons/ $km^2$) followed by West Bengal (1029 persons/ $km^2$) and Kerala (859 persons/ $km^2$). Among India's cities, Mumbai is the largest populous metropolitan city in India, in 2018 and it accommodates 22.1 million people, followed by Delhi with 28 million people and Kolkata.

**Table 1:** Rural and urban distribution of population and density of India/State/Union Territory: Census 2011 [38]

| Rank | State or Union territory | Population | Rural population (%) | Urban population (%) | Density (per $km^2$) |
|---|---|---|---|---|---|
| 1 | Uttar Pradesh | 199812341 | 77.73 | 22.27 | 828 |
| 2 | Maharashtra | 112374333 | 54.78 | 45.22 | 365 |
| 3 | Bihar | 104099452 | 88.71 | 11.29 | 1,102 |
| 4 | West Bengal | 91276115 | 68.13 | 31.87 | 1,029 |
| 5 | Madhya Pradesh | 72626809 | 72.37 | 27.63 | 236 |
| 6 | Tamil Nadu | 72147030 | 51.6 | 48.4 | 555 |
| 7 | Rajasthan | 68548437 | 75.13 | 24.87 | 201 |
| 8 | Karnataka | 61095297 | 61.33 | 38.67 | 319 |
| 9 | Gujarat | 60439692 | 57.4 | 42.6 | 308 |
| 10 | Andhra Pradesh | 49,577,103 | 70.53 | 29.47 | 303 |
| 11 | Odisha | 41974219 | 83.31 | 16.69 | 269 |
| 12 | Telangana | 35003674 | 61.12 | 38.88 | 312 |
| 13 | Kerala | 33406061 | 52.3 | 47.7 | 859 |
| 14 | Jharkhand | 32988134 | 75.95 | 24.05 | 414 |
| 15 | Assam | 31205576 | 85.9 | 14.1 | 397 |
| 16 | Punjab | 27743338 | 62.52 | 37.48 | 550 |
| 17 | Chhattisgarh | 25545198 | 76.76 | 23.24 | 189 |
| 18 | Haryana | 25351462 | 65.12 | 34.88 | 573 |
| 19 | Uttarakhand | 10086292 | 69.77 | 30.23 | 189 |
| 20 | Himachal Pradesh | 6864602 | 89.97 | 10.03 | 123 |
| 21 | Tripura | 3673917 | 73.83 | 26.17 | 350 |
| 22 | Meghalaya | 2966889 | 79.93 | 20.07 | 132 |
| 23 | Manipur | 2570390 | 69.79 | 30.21 | 122 |
| 24 | Nagaland | 1978502 | 71.14 | 28.86 | 119 |
| 25 | Goa | 1458545 | 37.83 | 62.17 | 394 |
| 26 | Arunachal Pradesh | 1383727 | 77.06 | 22.94 | 17 |
| 27 | Mizoram | 1097206 | 47.89 | 52.11 | 52 |
| 28 | Sikkim | 610577 | 74.85 | 25.15 | 86 |
| NCT | Delhi | 16787941 | 2.5 | 97.5 | 11,297 |
| UT1 | Jammu and Kashmir | 12267032 | 73.89 | 26.11 | 297 |

| UT2 | Puducherry | 1247953 | 31.67 | 68.33 | 2,598 |
| UT3 | Chandigarh | 1055450 | 2.75 | 97.25 | 9,252 |
| UT4 | Dadra and Nagar Haveli and Daman and Diu | 585764 | 41.57 | 58.43 | 970 |
| UT5 | Andaman and Nicobar Islands | 380581 | 62.3 | 37.7 | 46 |
| UT6 | Ladakh | 274000 | 16 | 84 | 2.8 |
| UT7 | Lakshadweep | 64473 | 21.93 | 78.07 | 2,013 |
| **Total** | **India** | **1210569573** | **68.84** | **31.16** | **382** |

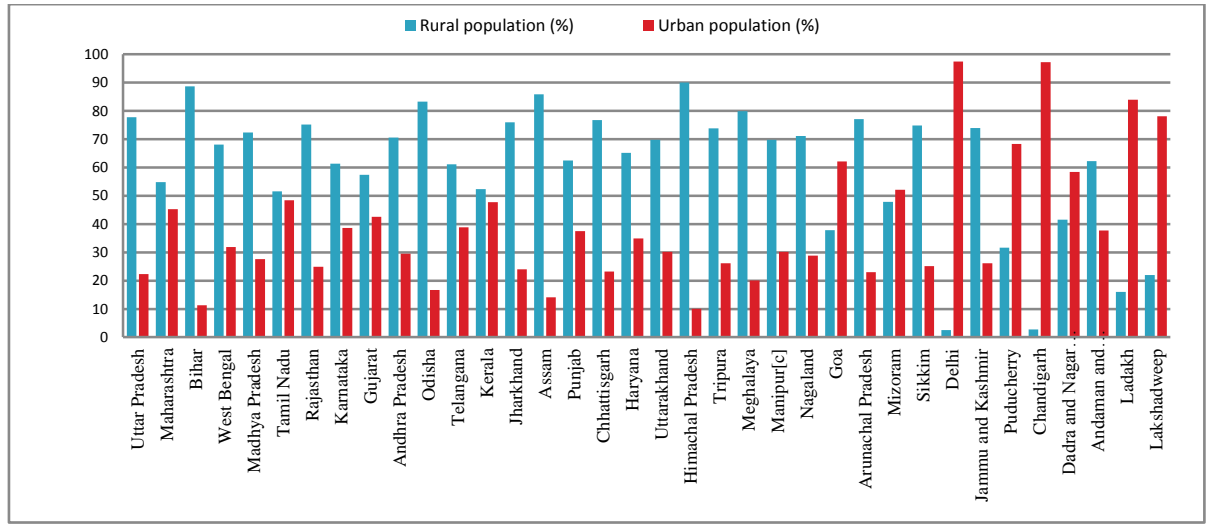

**Fig. 5:** Rural and urban distribution of the population of India/State/Union Territory: Census 2011 [39]

## 7. Validation and prediction of COVID-19 active and recovered cases

In this section, we demonstrate the experiments using the proposed model i.e. the SIR model with intervention on COVID-19 data. We have estimated the transmission rate $\beta$ from the prevalence of COVID-19 data. We also have validated as well as predicted the population of the infected and recovered population in our model for some of the most affected states of India. The experimental results being performed by using Python with panda [40], which is discussed in the following subsections:

### 7.1 Transmission rate ($\beta$) estimation from prevalence data

To find a more accurate value the transmission rate ($\beta$) for the SIR model with intervention is a challenging task for the COVID-19. In this study, we have estimated the value of $\beta$ from the prevalence reported COVID-19 cases in India for the duration of June 05, 2020, to July 25, 2020. Using Eq. (7), we have calculated the $\beta$ for 50 days and then calculated the $\beta$ using the linear regression and machine learning approaches. The *sklearn*, package of Python is used with a linear model and 10 cross-validations to obtain the optimal value of $\beta$. The estimated value of transmission rate ($\beta$) for India from June 05, 2020, to July 25, 2020, is shown in Fig. 6. Using this approach, the value of transmission rate ($\beta$) is estimated for India is 0.1738. The predicted values of the transmission rate ($\beta$) for some states of India are shown in Table 2.

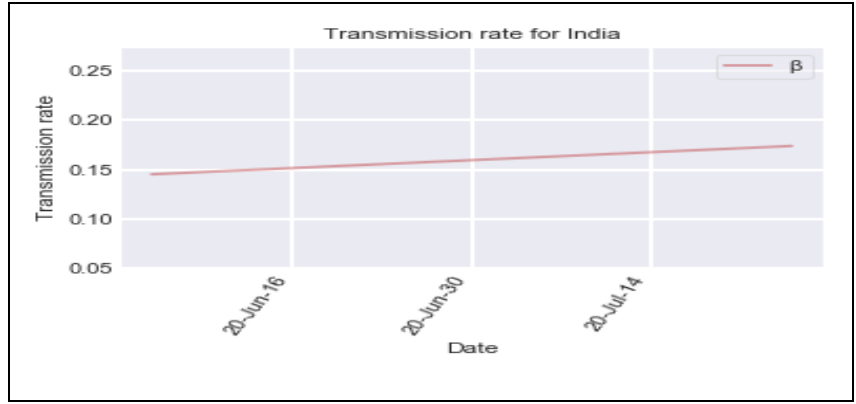

**Fig. 6**: Transmission rate ($\beta$) for India using linear regression

**Table 2:** Estimated transmission rate ($\beta$) using linear regression and intervention ($\rho$) value

| States/Country | Transmission rate ($\beta$) | Intervention value ($\rho$) |
|---|---|---|
| India | 0.1738 | 0.432 |
| Uttar Pradesh | 0.1908 | 0.445 |
| Maharashtra | 0.1443 | 0.338 |
| Tamil Nadu | 0.1713 | 0.460 |
| West Bengal | 0.2072 | 0.505 |
| Telangana | 0.2086 | 0.440 |
| Gujarat | 0.1556 | 0.420 |
| Bihar | 0.2666 | 0.610 |
| Arunachal Pradesh | 0.1815 | 0.310 |
| Assam | 0.2331 | 0.550 |

**Table 3:** Predicted COVID-19 active and recovered cases till Sept 30, 2020 with 95% and 99% confidence interval

| States/Country | Predicted COVID-19 active cases till Sept 30, 2020 | 95% confidence interval for predicted COVID-19 active cases | 99% confidence interval for predicted COVID-19 active cases | Predicted COVID-19 recovered cases till Sept 30, 2020 | 95% confidence interval for recovered COVID-19 cases | 99% confidence interval for recovered COVID-19 cases |
|---|---|---|---|---|---|---|
| India | 2850000 | [1145915, 1466626] | [1095527, 1517014] | 7220000 | [2678392, 3530126] | [2544575, 3663943] |
| Uttar Pradesh | 224367 | [73132, 99827] | [68938, 104021] | 460403 | [145182, 200735] | [136454, 209463] |
| Maharashtra | 684599 | [311399, 384900] | [299851, 396448] | 1890000 | [714533, 943126] | [678618, 979041] |
| Tamil Nadu | 206312 | [101135, 122179] | [97829, 125485] | 690605 | [320444, 393992] | [308889, 405548] |
| West Bengal | 154212 | [55267, 73253] | [52441, 76079] | 347641 | [118328, 159902] | [111796, 166434] |
| Telangana | 244048 | [61739, 91688] | [57034, 96393] | 415944 | [120293, 168496] | [112720, 176069] |
| Gujarat | 44656 | [23118, 27475] | [22434, 28160] | 161752 | [79075, 95766] | [76452, 98389] |
| Bihar | 101569 | [34885, 46840] | [33006, 48719] | 223043 | [75843, 102196] | [71702, 106336] |
| Arunachal Pradesh | 23425 | [5017, 7910] | [4562, 8364] | 32083 | [6286, 10276] | [5659, 10903] |
| Assam | 72292 | [24405, 32961] | [23060, 34305] | 162000 | [58657, 77114] | [55757, 80014] |

### 7.2 Validation and Prediction of COVID-19 active and recovered cases

In our analysis, simulations are carried out on COVID-19 data with taking active infected and recovered (including death cases) cases on July 25, 2020, as the initial value. Here we validated COVID-19 cases from June 05, 2020, to July 25, 2020, and further predict the COVID-19 cases from July 25, 2020, to Sept 30, 2020. And, also find the confidence interval [41] for the predicted COVID-19 cases. For the simulations, we have used the estimated values of the transmission rate ($\beta$) for India and some states of India. Since, on average it takes 14 days to recover from COVID-19 infection, so we have taken recovery rate $\gamma = 1/14$ in our experiments. In the coming subsections, our experimental results are discussed.

#### 7.2.1 India

For India, we find the transmission rate ($\beta$) and determine the optimal value of intervention $\rho = 0.432$ to fit the SIR model with intervention on COVID-19 prevalence data from June 05, 2020, to July 25, 2020, shown in Table 2. Fig. 7(a) shows the validation graph for actual and predicted active cases and Fig. 7(b) shows the validation of actual and predicted recovered COVID-19 cases from earlier COVID-19 cases. Predicted value by the SIR model with intervention is near the actual COVID-19 values for COVID-19 active cases and recovered cases. Due to diversity in India, different types of interventions have been implemented during the unlock phase of lockdown. In some states are still imposing full or partial lockdown and also depend on peoples living in the urban and rural areas. As 68.84% of the total population lives in the rural area (Table 1 and Fig. 5) which is generally isolated and mostly depends on the agriculture profession. Further, with the same parameters, predicted the number of infected and recovered cases for India till Sept 30, 2020, shown in Fig. 8 and depicted in Table 3. As per the present scenario, India will witness 28.5 lakhs COVID-19 active cases and 72.2 lakhs recovered cases till Sept 30, 2020. Also estimate the 95% and 99% confidence interval for the predicted and recovered COVID-19 cases, depicted in Table 3. The estimated 95% and 99% confidence interval for the predicted COVID-19 active cases are [1145915, 1466626] and [1095527, 1517014] respectively till Sept 30, 2020.

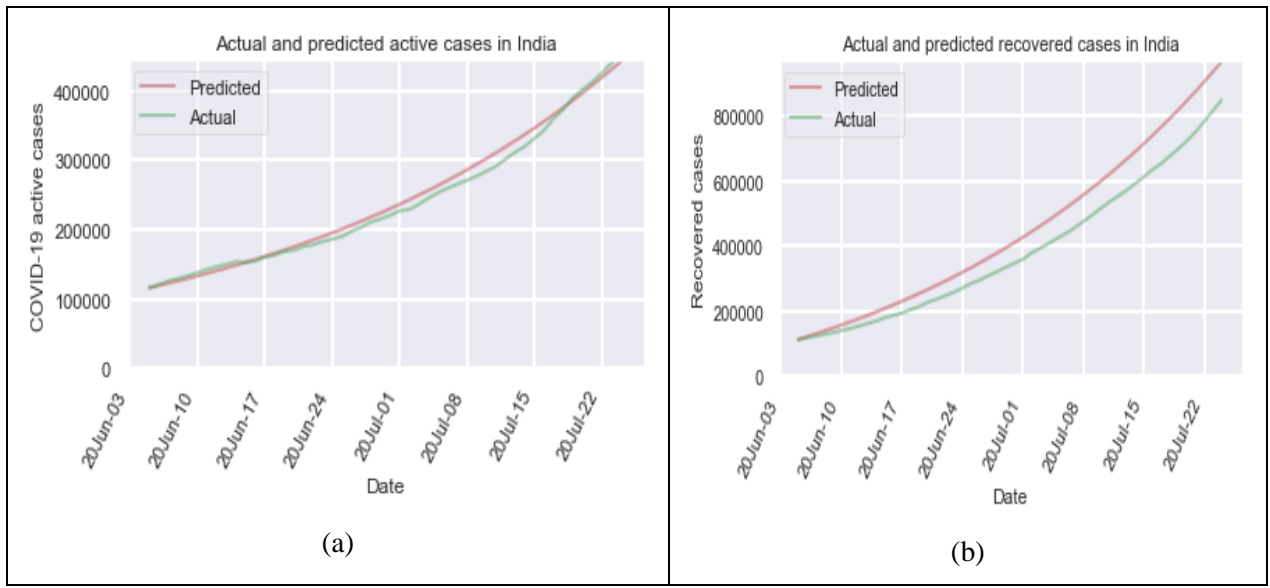

**Fig. 7:** Validation of predicted and actual COVID-19 active and recovered cases for India

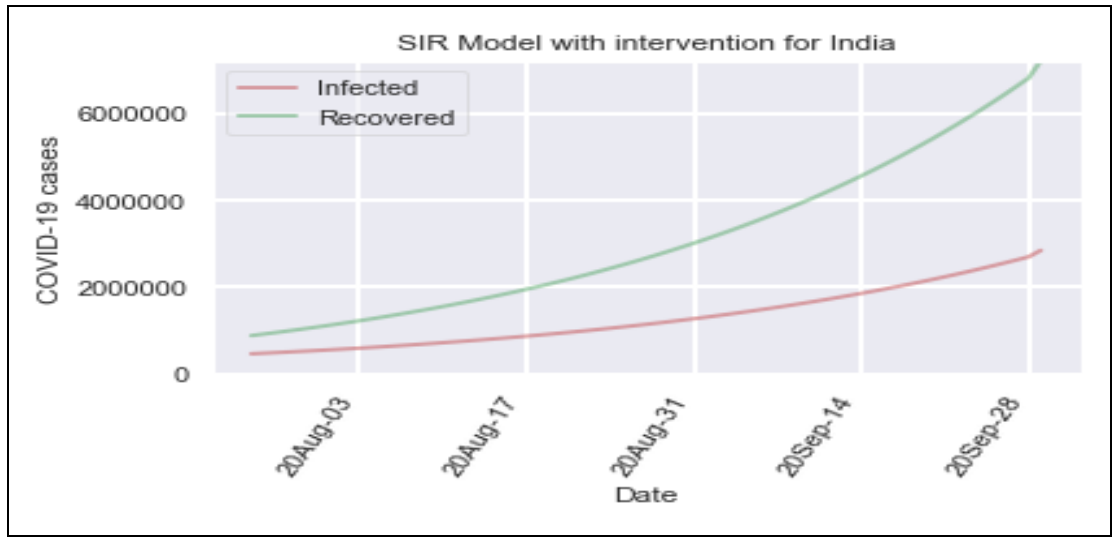

**Fig. 8:** Prediction of COVID-19 active and recovered cases for India

*7.2.2 Uttar Pradesh*

Uttar Pradesh is the most populous state in India with a density of 828 people per square kilometer and 77.73% of the populations are residing in the rural area according to census 2011, depicted in Table 1. This makes controlling the COVID-19 outbreak in Uttar Pradesh is a challenging task. The estimated transmission rate ($\beta$) for Uttar Pradesh is 0.1908 and we find the optimal intervention value $\rho(= 0.445)$ with fitting the actual COVID-19 cases from June 05, 2020, to July 25, 2020, shown in Table 2. Fig. 9(a) and Fig. 9(b) shows the validation of predicted and actual cases of COVID-19 active cases and recovered cases respectively, and it also shows that actual and predicted cases are in close agreement. Uttar Pradesh's government also initiated to spread awareness about COVID-19 among the people through posters and banners. Also during India's unlock phase, Uttar Pradesh government initiated the weekend (Saturday and Sunday) lockdown from July 11, 2020, to check the COVID-19 spread [42] as the complete lockdown affect the economy of the state. Further, we predict till Sept 30, 2020, of COVID-19 active and recovered cases with assuming the same parameters used in the validation, represented in Fig. 10 and in Table 3 along with 95% and 99% confidence interval estimation. On Sept 30, 2020, the number of COVID-19 active cases will be about 2.24 lakhs, and recovered cases will be around 4.60 lakhs in Uttar Pradesh. Table 3, shows that 95% and 99% confidence interval for the predicted COVID-19 active cases are [73132, 99827] and [68938, 104021] respectively for Uttar Pradesh till Sept 30, 2020.

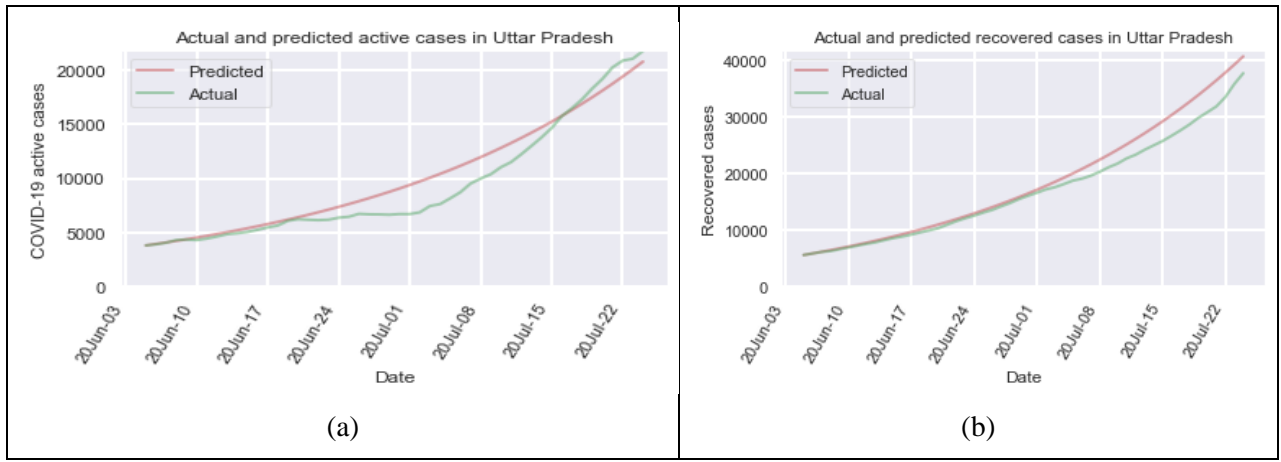

**Fig. 9:** Validation of predicted and actual COVID-19 active and recovered cases for Uttar Pradesh

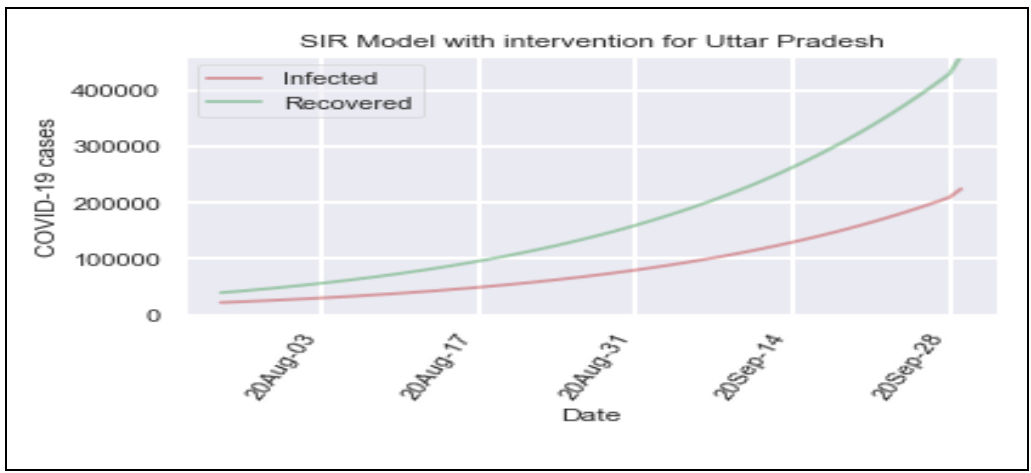

**Fig. 10:** Prediction of COVID-19 active and predicted cases for Uttar Pradesh

### 7.2.3 Maharashtra

Maharashtra is the second most populous state in India with a population density of 365 people per square kilometer and 45.22% of population is residing in the urban area according to census 2011. Maharashtra accounts for nearly one-third of the total COVID-19 cases in India with about 40% of all deaths [43]. Mumbai in Maharashtra is the largest metropolis and the most populous city, also worst-affected with COVID-19 city in India. Because of rising COVID-19 cases, during India's unlocking phase, the Maharashtra government has extended the lockdown with some relaxation in the entire state till Aug 31, 2020 [44, 45]. In the case of Maharashtra, we estimated the transmission rate $\beta(=0.1443)$, and with fitting the actual COVID-19 cases from June 05, 2020, to July 25, 2020, we determine the intervention value $\rho(=0.338)$, shown in Table 2. The validation of actual and predicted cases using the SIR model with intervention for active COVID-19 and recovered cases is shown in Fig. 11(a) and Fig. 11(b) respectively. The actual and predicted results are almost similar but for recovered cases, the actual and predicted case shows a slight high variation. Here, the predicted recovered cases also include the deceased case whereas the actual recovered cases considered only recovered cases. A further prediction is made until Sept 30, 2020, with the same parameters which are shown in Fig. 12 and Table 3. The confidence intervals for active and recovered cases are estimated and shown in Table 3. On Sept 30, 2020, COVID-19 active and recovered cases will be less than 6.84 lakhs and 18.9 lakhs respectively in Maharashtra. The estimated 95% and 99% confidence interval for predicted COVID-19 active cases are [311399, 384900] and [299851, 396448] respectively.

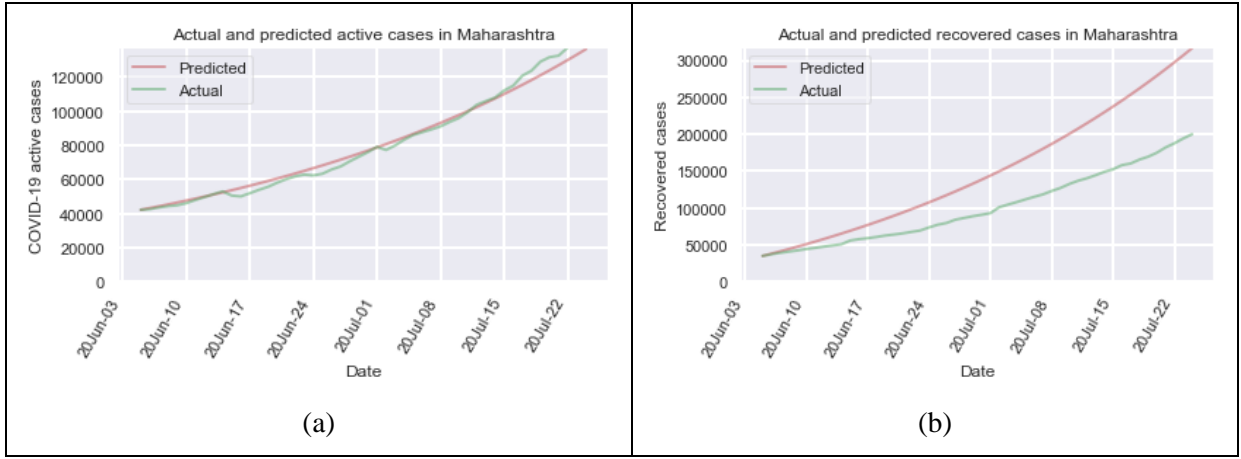

(a)          (b)

**Fig. 11:** Validation of predicted and actual COVID-19 active and recovered cases for Maharashtra

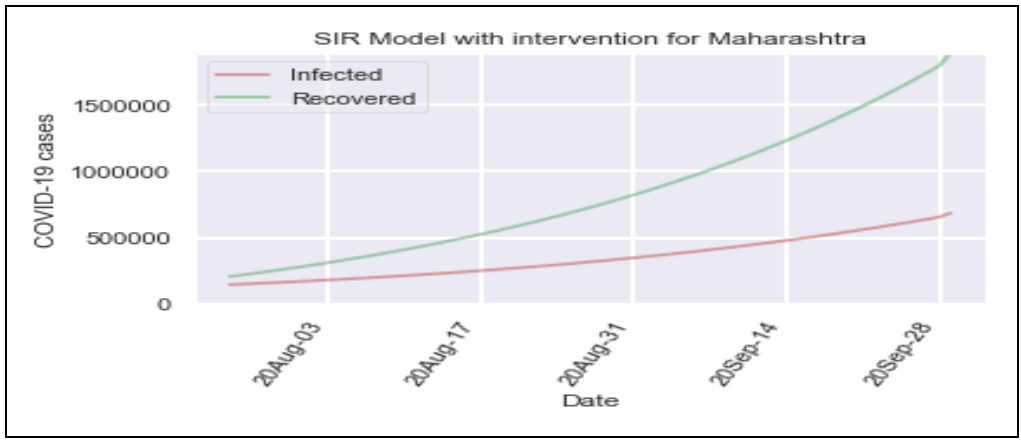

**Fig. 12:** Prediction of COVID-19 active and recovered cases for Maharashtra

### 7.2.4 Tamil Nadu

Tamil Nadu is the sixth most populous state in India and 48.4% of people are living in the urban areas which are more crowded than the rural area, and managing the social distance is difficult in comparison to the rural population. Chennai, the capital city of Tamil Nadu is the fourth largest populous metropolitan in India. Tamil Nadu has recorded the second-highest number of cases in India and Chennai city being the worst affected [37]. More than half of the confirmed cases in Tamil Nadu are from Chennai. To curb the COVID-19 spread, the Tamil Nadu government has extended the lockdown with some relaxation in the entire state till Aug 31, 2020 [46]. For Tamil Nadu, we estimated the transmission rate $\beta (= 0.1713)$ from the prevalence COVID-19 cases from June 25, 2020, to July 25, 2020, and determined the intervention value $\rho(= 0.460)$ to fit the SIR model with intervention, and depicted in Table 2. Fig. 13(a) and Fig. 13(b) show the validation graph of predicted and actual cases for active cases and recovered cases. Till July 17, 2020, the actual cases are more than the predicted cases. With the same parameter setting, we did the prediction for Tamil Nadu, shown in Fig. 14 and depicted in Table 3. At the current rate of growth, Tamil Nadu will witness 2.06 lakhs active COVID-19 cases and 6.9 lakh recovered cases till Sept 30, 2020. The estimated confidence intervals for predicted active and recovered cases at level 95% and 99% are also tabulated in Table 3.

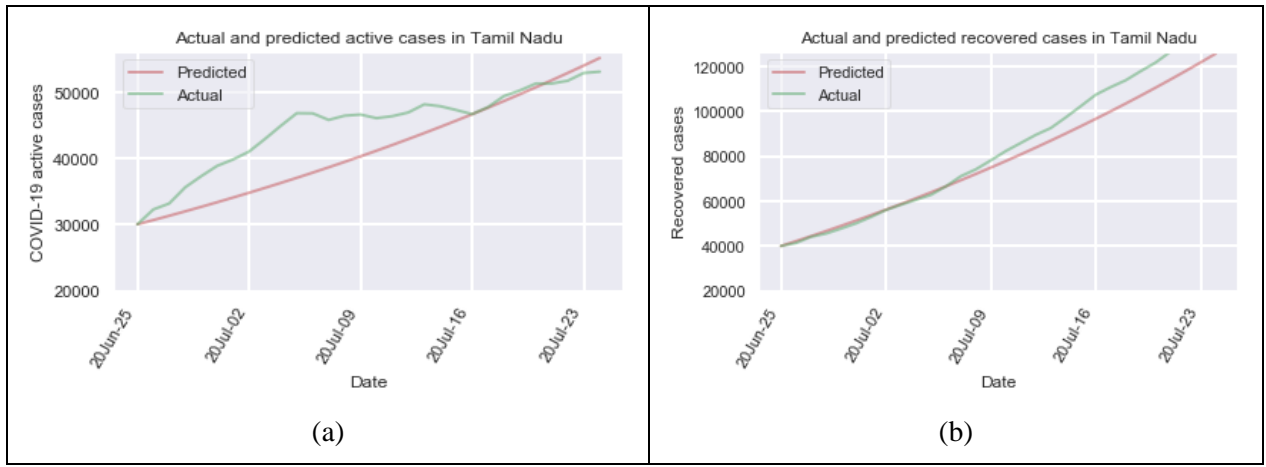

**Fig. 13:** Validation of predicted and actual COVID-19 active and recovered cases for Tamil Nadu

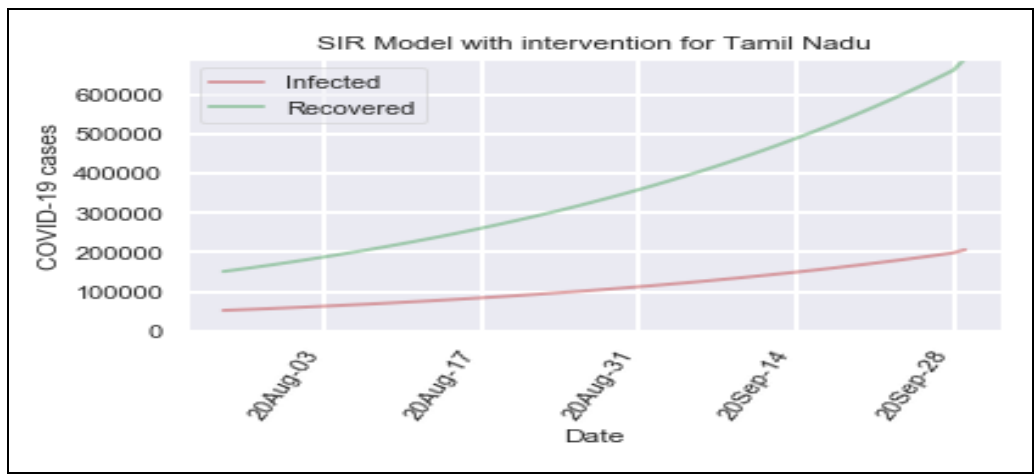

**Fig. 14:** Prediction of COVID-19 active and recovered cases for Tamil Nadu

### 7.2.5 West Bengal

West Bengal is the fourth most populous state in India with a population density of 1029 per square kilometer (Table 1) and Kolkata is the third most populous city in India. 68.13 % of people of West Bengal are residing in rural areas that are less crowded compared to the urban area. To combat the COVID-19, the West Bengal government has implemented the complete lockdown in July 2020 and selected day's complete lockdown in August 2020 with some relaxation in the entire state [47, 48]. We estimate the transmission rate $\beta(= 0.2072)$ for West Bengal from prevalence data from June 05, 2020, to July 25, 2020, and determine the intervention value $\rho(= 0.505)$ with fitting the actual cases by the SIR model with intervention, display in Table 2. The validation of actual and predicted cases using the SIR model with intervention for active COVID-19 and recovered cases are shown in Fig. 15(a) and Fig. 15(b) respectively. Prediction for West Bengal is shown in Fig. 16 and depicted in Table 3 with the same parameters. At the current rate of growth, West Bengal will witness 1.54 lakh active COVID-19 cases and 3.47 lakh recovered cases till Sept 30, 2020. For predicted active and recovered COVID-19 cases, the confidence interval is also estimated, shown in Table 3. The estimated 95% and 99% confidence interval for predicted COVID-19 active cases are [55267, 73253] and [52441, 76079] respectively.

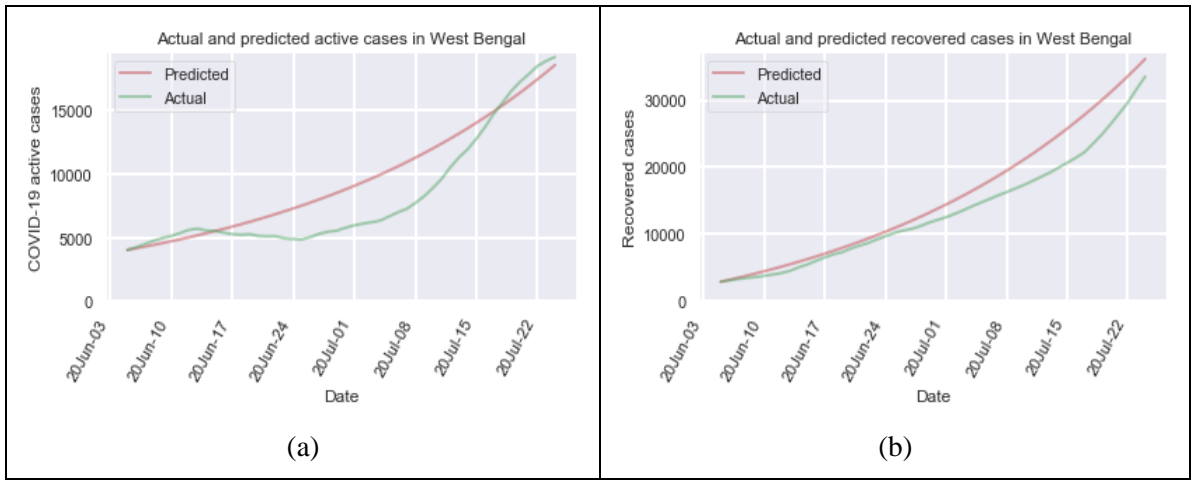

**Fig. 15:** Validation of predicted and actual COVID-19 active and recovered cases for West Bengal

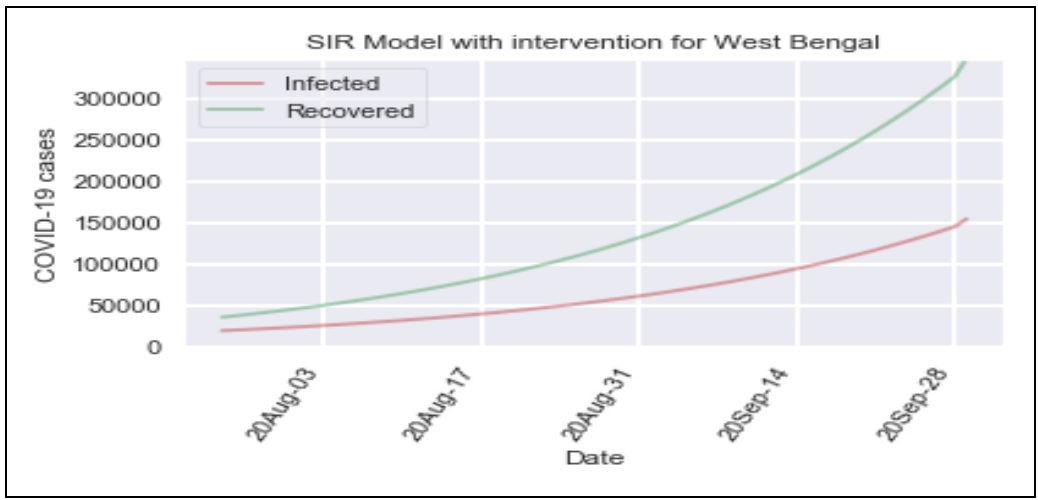

**Fig. 16:** Prediction of COVID-19 active and recovered cases for West Bengal

### 7.2.6 Telangana

In Telangana, 61.12% of people are residing in the rural area and population density is 312 per square kilometer (Table 1). From the prevalence COVID-19 data from June 05, 2020, to July 25, 2020, we estimate the transmission rate $\beta(= 0.2086)$ and further determine the intervention value $\rho(= 0.440)$ with fitting the SIR model with intervention with the actual COVID-19 cases, shown in Table 2. Fig. 17(a) and Fig. 17(b) represent the validation graph of the predicted and actual COVID-19 cases. The predicted and actual cases show that variation because of different interventions implemented during the period. With the same parameter discussed above, the prediction has made till Sept 30, 2020, shown in Fig. 18 and represented in Table 3. With the present growth scenario, Telangana will witness 2.44 lakhs COVID-19 active cases and 4.15 lakhs recovered cases till Sept 30, 2020. We have also estimated confidence intervals for predicted active and recovered cases at level 95% and 99% for Telangana, which are shown in Table 3.

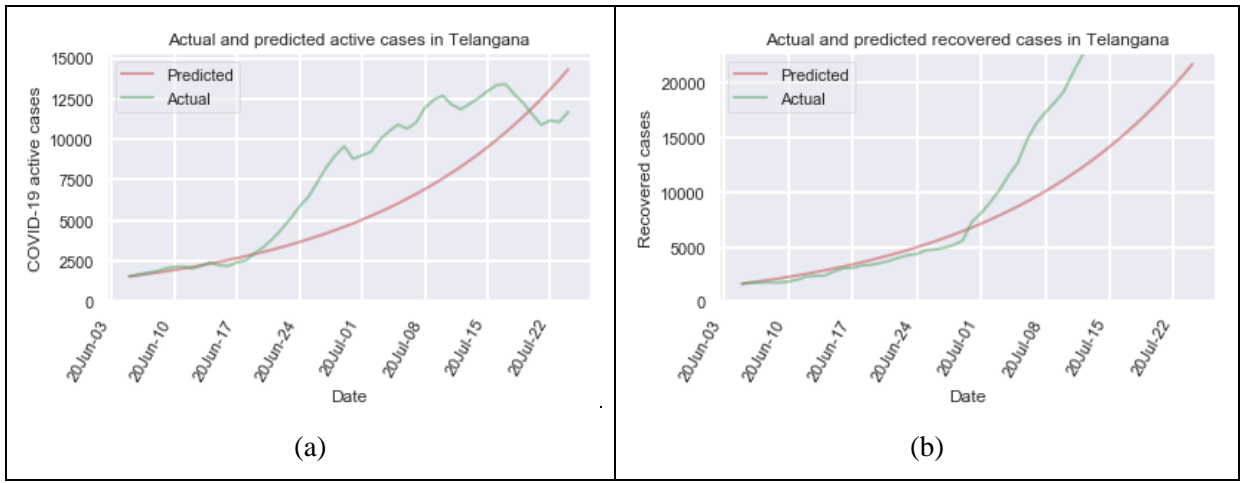

**Fig. 17:** Validation of predicted and actual COVID-19 active and recovered cases for Telangana

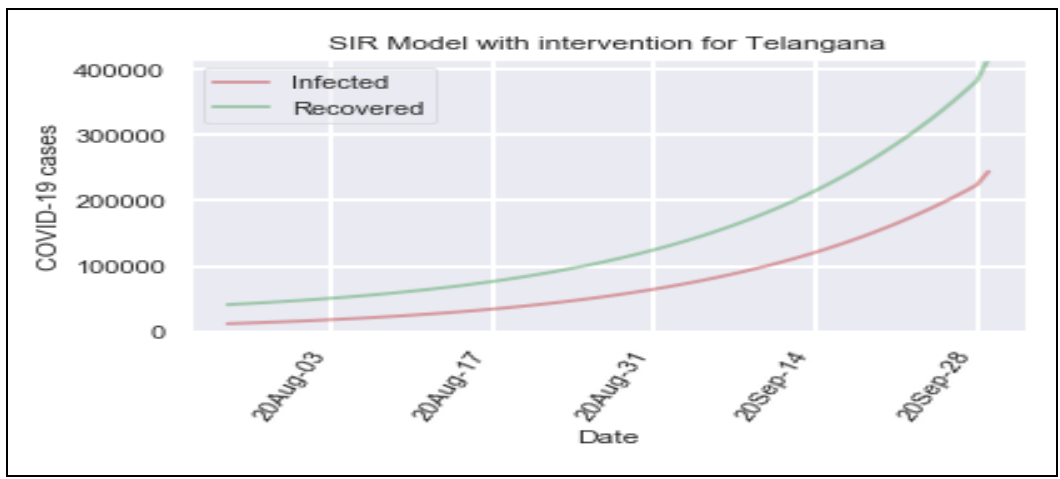

**Fig. 18:** Prediction of COVID-19 active and recovered cases for Telangana

### 7.2.7 Gujarat

In Gujarat, 57.4 % population is residing in rural areas with a population density of 308 people per square kilometer (Table 1 and Fig. 5). The rural area is more separated in comparison to the urban area and also population density is low. To combating from the COVID-19, isolation is a key approach and also depends on people's behavior. For Gujarat, we estimated the transmission rate $\beta(= 0.1556)$ and determined the intervention value $\rho(= 0.420)$ with fitting the predicted COVID-19 cases by the SIR model with intervention with the actual COVID-19 cases from the prevalence COVID-19 data during June 05, 2020, to July 25, 2020, and shown in Table 2. Fig. 19(a) and Fig. 19(b) represent the validation graph of the predicted and actual COVID-19 cases for active and recovered cases respectively from prevalence data. With the same parameter discussed above, the prediction has made till Sept 30, 2020, shown in Fig. 20 and also depicted in Table 3. Gujarat will have about 44656 COVID-19 active cases and 1.61 lakhs recovered cases till Sept 30, 2020, if the current situation prevails. Also estimated 95% and 99%, confidence interval for predicted active and recovered cases for Gujarat, depicted in Table 3. The estimated 95% and 99% confidence interval for predicted COVID-19 active cases are [23118, 27475] and [22434, 28160] respectively.

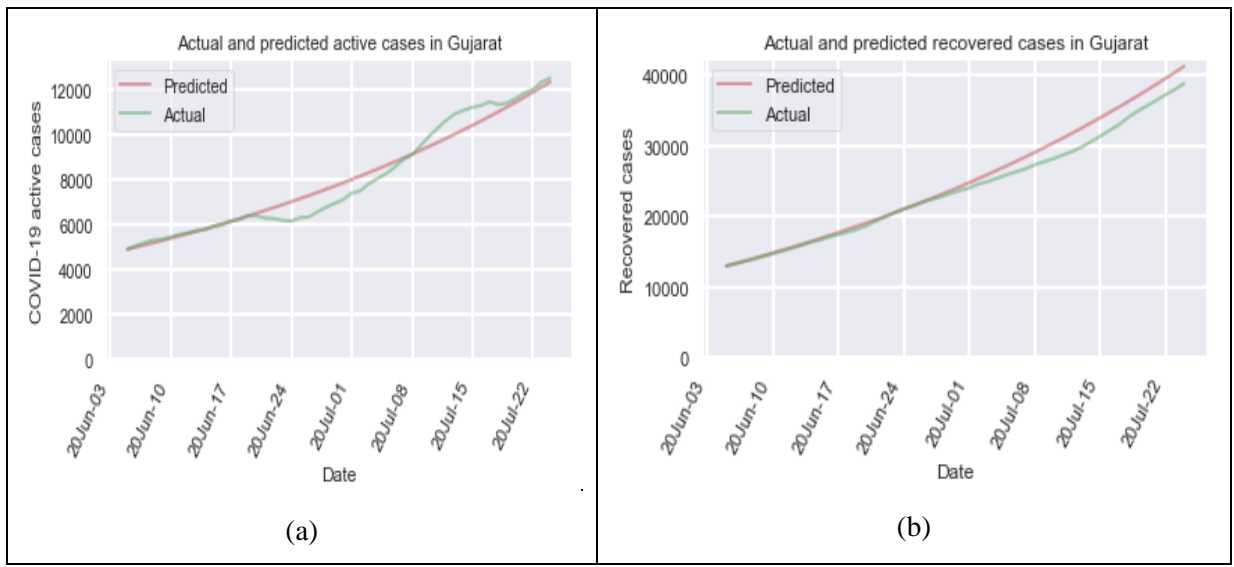

**Fig. 19:** Validation of predicted and actual COVID-19 active and recovered cases for Gujarat

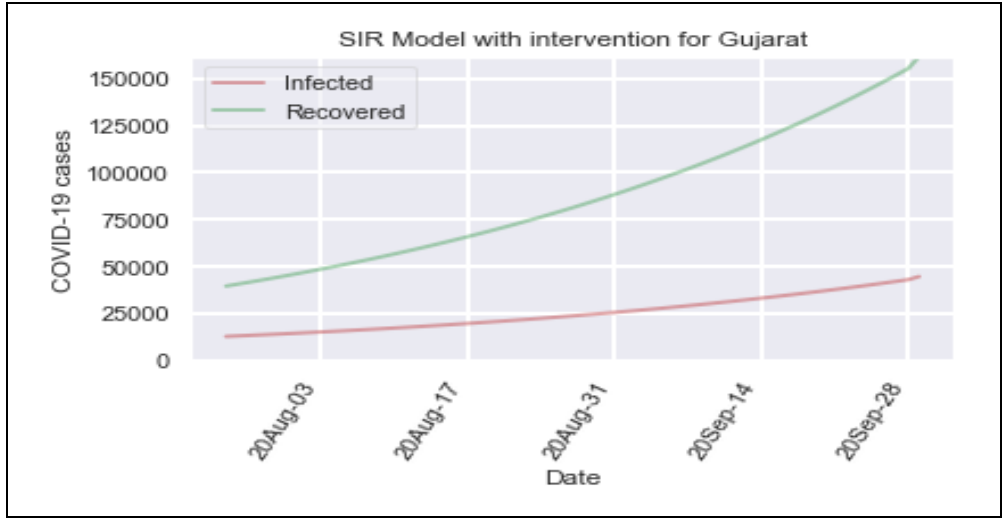

**Fig. 20:** Prediction of COVID-19 active and recovered cases for Gujarat

### *7.2.8 Bihar*

According to the 2011 India census (Table 1), Bihar is the third most populous and most densely populated state of India with 1,106 persons per square kilometer. 88.71 % population of Bihar is residing in the rural area and it is the second lowest urbanized state in India. The major proportion of the population in Bihar resides in the villages which is more isolated. Bihar is under complete lockdown since Mar 2020 with some relaxation [49]. Bihar has the largest population of migrant laborers in India. During the national lockdown, a lot of migrant workers returned their homes in Bihar as a result of the shutdown of the industry and losing their jobs. In this situation, the Bihar government made the compulsory quarantine to returning migrants. Under the complete lockdown scenario, we estimate the transmission rate $\beta (= 0.2666)$ and determine the intervention value $\rho (= 0.610)$ with fitting the predicted COVID-19 cases by the SIR model with intervention with the actual COVID-19 cases from the prevalence COVID-19 data during June 05, 2020, to July 25, 2020, shown in Table 2. Due to completer lockdown, the intervention value is more for Bihar. Fig. 21(a) and Fig. 21(b) show the validation graph of the predicted and actual COVID-19 cases for active and recovered cases respectively from prevalence data. There is variation is the predicted and actual value which depends on the intervention imposed by the government and individual behaviors. With the same parameter discussed above, the prediction has made till Sept 30, 2020, shown in Fig. 22 and also depicted in Table 3. If the present COVID-19 situation remains unchanged, Bihar will witness about 1.01 lakhs COVID-19 active cases and 2.23 lakhs recovered cases till Sept 30, 2020. Also estimated 95% and 99% confidence intervals for predicted active and recovered cases for Bihar are depicted in Table 3. The confidence interval are [34885, 46840] and [33006, 48719] for predicted active cases and [75843, 102196] and [71702, 106336] for the recovered cases at 95% and 99% level respectively.

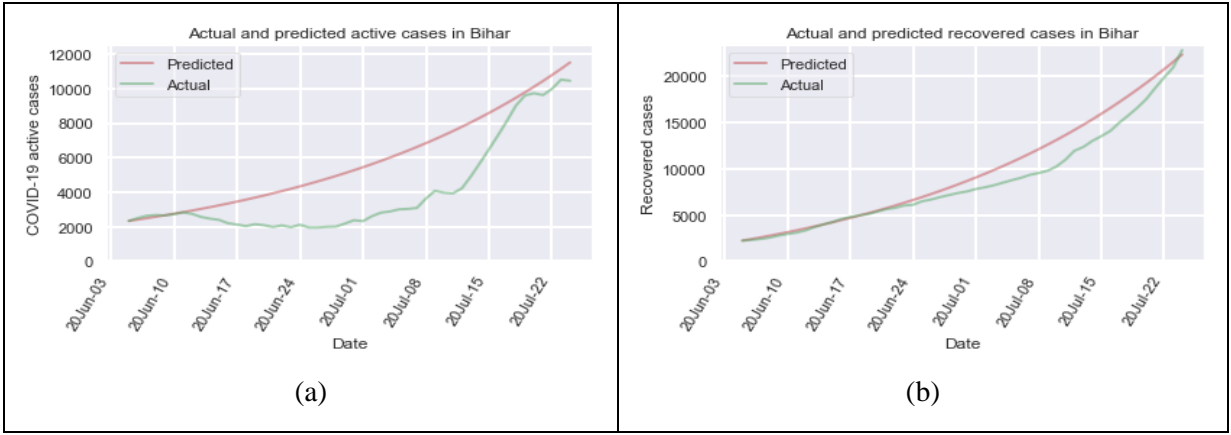

**Fig. 21:** Validation of predicted and actual COVID-19 active and recovered cases for Bihar

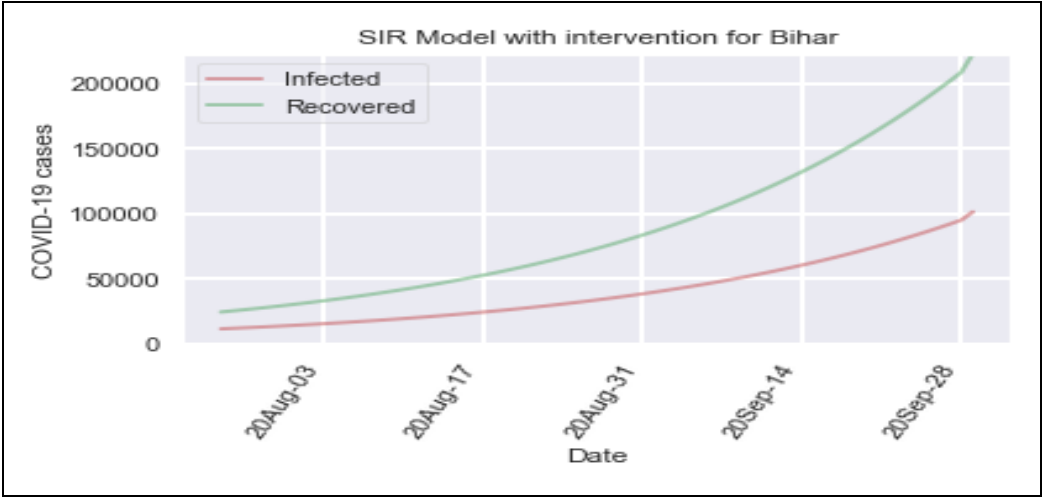

**Fig. 22:** Prediction of COVID-19 active and recovered cases for Bihar

### 7.2.9 *Arunachal Pradesh*

Arunachal Pradesh is a northeast state, where nearly 77% population are living in the rural area and population density is about 17 people per square kilometer, second lowest population density state, according to India census 2011 (Table 1). It is mostly isolated as being the more residing in the rural area and population density is less. For Arunachal Pradesh, we estimate the transmission rate $\beta(= 0.1815)$ and determine the intervention value $\rho(= 0.310)$ with fitting the predicted COVID-19 cases by SIR model with intervention with the actual COVID-19 cases from the prevalence COVID-19 data during June 05, 2020, to July 25, 2020, depicted in Table 2. Fig. 23(a) and Fig. 23(b) show the validation graph of the predicted and actual COVID-19 cases for active and recovered cases respectively from prevalence data. Predicted recovered cases show more than the actual cases. With the same parameter discussed above, the prediction has made till Sept 30, 2020, shown in Fig. 24 and also depicted in Table 3. In the present COVID-19 growth scenario, Arunachal Pradesh will witness not more than 23425 COVID-19 active cases and 32083 recovered cases till Sept 30, 2020. Also estimated 95% and 99% confidence interval for predicted active and recovered cases for Arunachal Pradesh depicted in Table 3. The estimated 95% and 99% confidence interval for predicted COVID-19 active cases are [5017, 7910] and [4562, 8364] respectively.

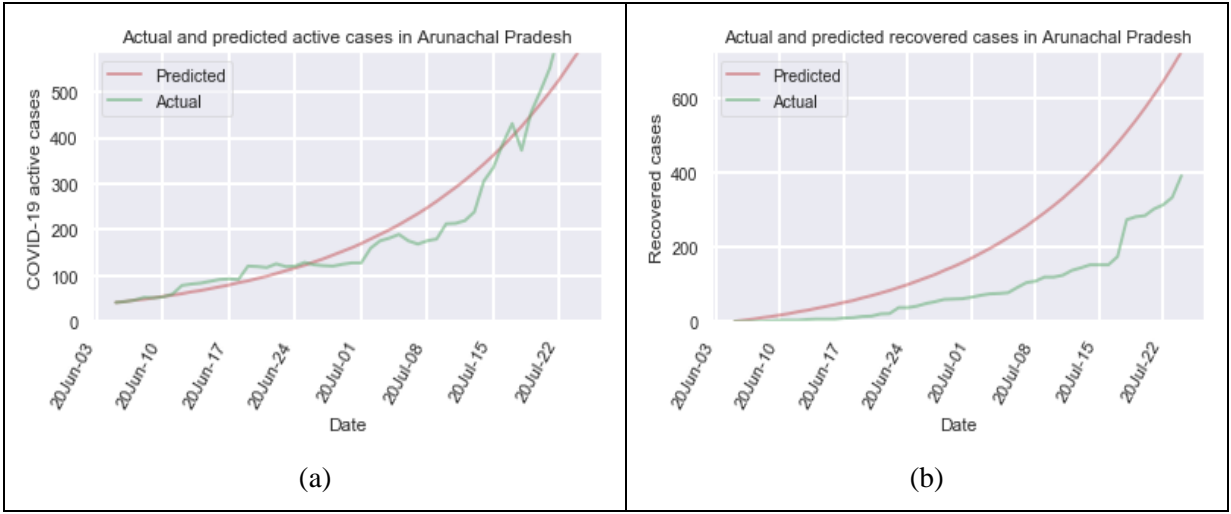

(a)          (b)

**Fig. 23:** Validation of predicted and actual COVID-19 active and recovered cases for Arunachal Pradesh

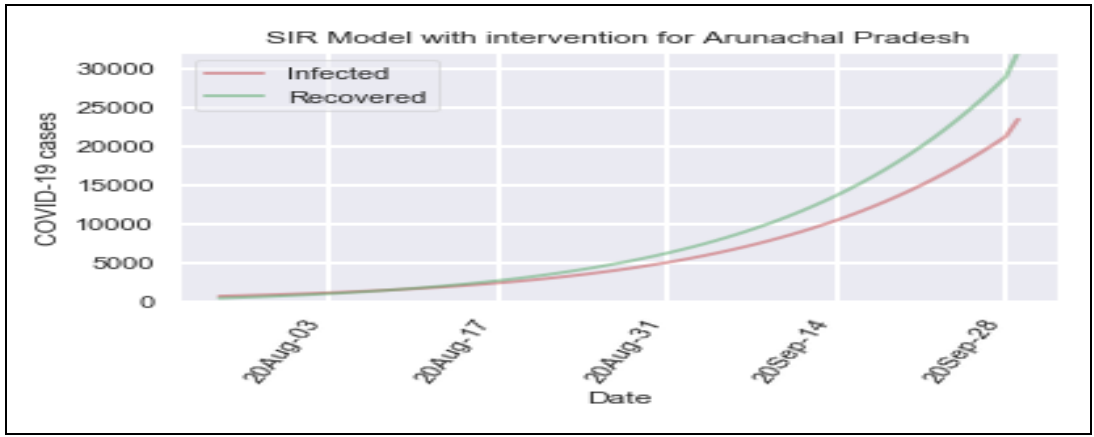

**Fig. 24:** Prediction of COVID-19 active and recovered cases for Arunachal Pradesh

### 7.2.10 Assam

As per the 2011 census, nearly 86% population of Assam is living in the rural area and the population density of Assam is 397 people per square kilometer (Table 1). Assam government imposed the complete lockdown in Guwahati from June 28, 2020, July 12, 2020, due to the rapid increase in the number of COVID-19 cases in the city [50]. The government also enforced weekend lockdown from June 27, 2020, in all towns. Guwahati city has been worse affected by COVID-19. We estimated the transmission rate $\beta(= 0.2331)$ and determine the intervention value $\rho(= 0.550)$ with fitting the predicted COVID-19 cases by SIR model with intervention with the actual COVID-19 cases from the prevalence COVID-19 data during June 05, 2020, to July 25, 2020. Fig. 25(a) and Fig. 25(b) show the validation graph of the predicted and actual COVID-19 cases for active and recovered cases respectively from prevalence data. In our prediction till Sept 30, 2020, shown in Fig. 26 and also depicted in Table 3. Our model predicts near about 72292 COVID-19 active cases and 1.62 lakhs recovered cases till Sept 30, 2020, in Assam.

Further estimated 95% and 99%, confidence interval for predicted active and recovered cases for Assam, depicted in Table 3.

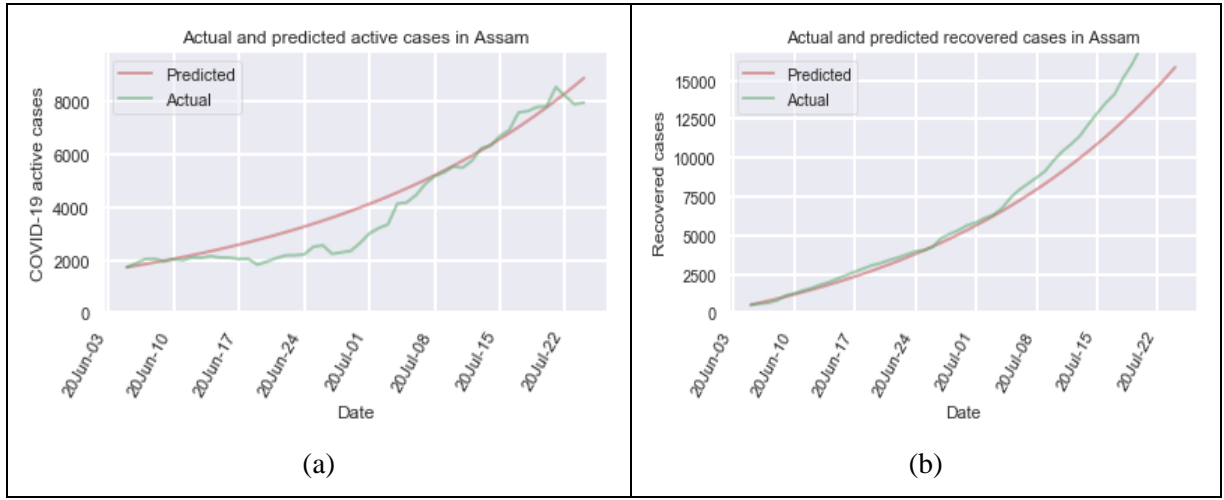

**Fig. 25:** Validation of predicted and actual COVID-19 active and recovered cases for Assam

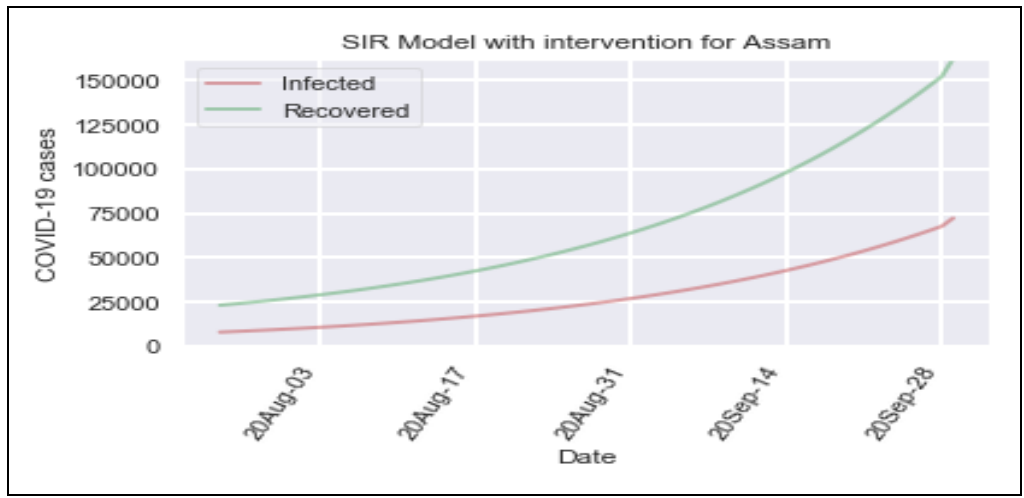

**Fig. 26:** Prediction of COVID-19 active and recovered cases for Assam

## 8. Conclusion

At present, the COVID-19 outbreak is a potential threat due to its rapid spread behavior. It is more threatening in India because the health facilities in India are low and India is a very densely populated country. Among vaccination, herd immunity, plasma therapy, and epidemiological interventions to combat with COVID-19, India has the only epidemiological interventions as the best possible option until the vaccine become available. Keeping this in mind, we incorporate the intervention in the classical SIR model and suggested the SIR model with intervention in this work. The supervised machine learning approach to estimate the transmission rate ($\beta$) for the SIR model with intervention from the prevalence of COVID-19 data in India and some states of India is also discussed. The work is implemented in Python and the prediction of COVID-19 active and recovered cases with validation with the optimal value of intervention for entire India and some states of India till Sept 30, 2020, have been made. We conclude our analysis in the following points.

- The imposed intervention is implemented to flatting the curve for COVID-19 spread in India.
- The epidemiological intervention especially isolation and social distancing, and other preventive measures are very effective to combat COVID-19 until the vaccine becomes available.
- There is a need for acceleration in tracing, testing, quarantining, isolating, and treating infected persons with medical advice at the early stage of COVID-19 infected especially in the rural areas. This will increase the value of $\rho$ which will reduce the effective transmission rate ($\beta(1-\rho)$) resulting in a reduced infected population.
- The predicted COVID-19 active and recovered cases till Sept 30, 2020, for entire India and states: Uttar Pradesh, Maharashtra, Tamil Nadu, West Bengal, Telangana, Gujarat, Bihar, Arunachal Pradesh, and Assam are depicted in Table 3.
- Estimated the confidence intervals at level 95% and 99% for the predicted active and recovered COVID-19 cases till Sept 30, 2020, depicted in Table 3.

## References

1. Dicker, R. C., Coronado, F., Koo, D., & Parrish, R. G. (2006). Principles of epidemiology in public health practice; an introduction to applied epidemiology and biostatistics.

2. Huang, C., Wang, Y., Li, X., Ren, L., Zhao, J., Hu, Y., ... & Cheng, Z. (2020). Clinical features of patients infected with 2019 novel coronavirus in Wuhan, China. *The lancet*, *395*(10223), 497-506.

3. WHO, naming the coronavirus disease (COVID-19) and the virus that causes it. Available online: https://www.who.int/emergencies/diseases/novel-coronavirus-2019/technical-guidance/naming-the-coronavirus-disease-(covid-2019)-and-the-virus-that-causes-it/

4. WHO Director-General's opening remarks at the media briefing on COVID-19, 11 March 2020. Available online: https://www.who.int/dg/speeches/detail/who-director-general-s-opening-remarks-at-the-media-briefing-on-covid-19---11-march-2020/

5. Transmission of SARS-CoV-2: implications for infection prevention precautions: https://www.who.int/news-room/commentaries/detail/transmission-of-sars-cov-2-implications-for-infection-prevention-precautions/

6. India confirms its first coronavirus case (Jan 30, 2020). Available online: https://www.cnbc.com/2020/01/30/india-confirms-first-case-of-the-coronavirus.html/

7. Ferguson, N. M., Laydon, D., Nedjati-Gilani, G., Imai, N., Ainslie, K., Baguelin, M., ... & Dighe, A. (2020). Impact of non-pharmaceutical interventions (NPIs) to reduce COVID-19 mortality and healthcare demand. 2020. *DOI*, *10*, 77482.

8. Wallinga, J., & Lipsitch, M. (2007). How generation intervals shape the relationship between growth rates and reproductive numbers. *Proceedings of the Royal Society B: Biological Sciences*, *274*(1609), 599-604.

9. Syal, K. (2020). COVID-19: herd immunity and convalescent plasma transfer therapy. *Journal of Medical Virology*.

10. Kwok, K. O., Lai, F., Wei, W. I., Wong, S. Y. S., & Tang, J. W. (2020). Herd immunity–estimating the level required to halt the COVID-19 epidemics in affected countries. *Journal of Infection*, *80*(6), e32-e33.

11. Randolph, H. E., & Barreiro, L. B. (2020). Herd Immunity: Understanding COVID-19. *Immunity*, *52*(5), 737-741.

12. WHO, Q&A on coronaviruses (COVID-19). Available online: https://www.who.int/emergencies/diseases/novel-coronavirus-2019/question-and-answers-hub/q-a-detail/q-a-coronaviruses/

13. Anderson, R. M., Heesterbeek, H., Klinkenberg, D., & Hollingsworth, T. D. (2020). How will country-based mitigation measures influence the course of the COVID-19 epidemic?. *The Lancet*, *395*(10228), 931-934.

14. India coronavirus: Modi announces 21-day nationwide lockdown, limiting movement of 1.4bn people (Mar 24, 2020). Available online: https://www.independent.co.uk/news/world/asia/india-coronavirus-lockdown-modi-speech-cases-update-news-a9421491.html/

15. More Weeks Of Lockdown Starting May 4". *NDTV.com*. Retrieved 1 May 2020. https://www.ndtv.com/india-news/nationwide-lockdown-over-coronavirus-extended-for-two-weeks-beyond-may-4-2221782/

16. Centre extends nationwide lockdown till May 31, new guidelines issued". *Tribuneindia News Service. 17 May 2020. Retrieved* 17 May *2020.*https://www.tribuneindia.com/news/nation/centre-extends-nationwide-lockdown-till-may-31-new-guidelines-issued-86042/

17. COVID-19 pandemic lockdown in India. https://en.wikipedia.org/wiki/COVID-19_pandemic_lockdown_in_India#cite_note-12/

18. Calafiore, G. C., Novara, C., & Possieri, C. (2020). A modified SIR model for the COVID-19 contagion in Italy. *arXiv preprint arXiv:2003.14391*.

19. Turner Jr, M. E., Bradley Jr, E. L., Kirk, K. A., & Pruitt, K. M. (1976). A theory of growth. *Mathematical Biosciences*, *29*(3-4), 367-373.

20. Chowell, G. (2017). Fitting dynamic models to epidemic outbreaks with quantified uncertainty: a primer for parameter uncertainty, identifiability, and forecasts. *Infectious Disease Modelling*, *2*(3), 379-398.

21. Richards, F. J. (1959). A flexible growth functions for empirical use. *Journal of experimental Botany*, *10*(2), 290-301.

22. Chowell, G., Tariq, A., & Hyman, J. M. (2019). A novel sub-epidemic modeling framework for short-term forecasting epidemic waves. *BMC medicine*, *17*(1), 164.

23. Kermack, W. O., & McKendrick, A. G. (1927). A contribution to the mathematical theory of epidemics. *Proceedings of the royal society of london. Series A, Containing papers of a mathematical and physical character*, *115*(772), 700-721.

24. Das, S. (2020). Prediction of COVID-19 Disease Progression in India: Under the Effect of National Lockdown. *arXiv preprint arXiv:2004.03147*.

25. Ndiaye, B. M., Tendeng, L., & Seck, D. (2020). Analysis of the COVID-19 pandemic by SIR model and machine learning technics for forecasting. *arXiv preprint arXiv:2004.01574*.

26. Ndiaye, B. M., Tendeng, L., & Seck, D. (2020). Comparative prediction of confirmed cases with COVID-19 pandemic by machine learning, deterministic and stochastic SIR models. *arXiv preprint arXiv:2004.13489*.

27. Liu, Z., Magal, P., Seydi, O., & Webb, G. (2020). Predicting the cumulative number of cases for the COVID-19 epidemic in China from early data. *arXiv preprint arXiv:2002.12298*.

28. Chen, Y. C., Lu, P. E., Chang, C. S., & Liu, T. H. (2020). A Time-dependent SIR model for COVID-19 with undetectable infected persons. *arXiv preprint arXiv:2003.00122*.

29. Deo, V., Chetiya, A. R., Deka, B., & Grover, G. Forecasting Transmission Dynamics of COVID-19 in India Under Containment Measures-A Time-Dependent State-Space SIR Approach.

30. Kobayashi, G., Sugasawa, S., Tamae, H., & Ozu, T. (2020). Predicting intervention effect for COVID-19 in Japan: state space modeling approach. *BioScience Trends*.

31. Adwibowo, A. (2020). Flattening the COVID 19 curve in susceptible forest indigenous tribes using SIR model. *medRxiv*.

32. Jennifer Ciarochi (2020), How COVID-19 and other Infectious Diseases Spread: Mathematical Modeling: https://triplebyte.com/blog/modeling-infectious-diseases/

33. Montgomery, D. C., & Runger, G. C. (2010). *Applied statistics and probability for engineers*. John Wiley & Sons.

34. Pollicott, M., Wang, H., & Weiss, H. (2012). Extracting the time-dependent transmission rate from infection data via solution of an inverse ODE problem. *Journal of biological dynamics*, *6*(2), 509-523.

35. Marinov, T. T., Marinova, R. S., Omojola, J., & Jackson, M. (2014). Inverse problem for coefficient identification in SIR epidemic models. *Computers & Mathematics with Applications*, *67*(12), 2218-2227.

36. Marinov, T. T., & Marinova, R. S. (2020). COVID-19 Analysis Using Inverse Problem for Coefficient Identification in SIR Epidemic Models.

37. Covid19India is a crowd-sourced open database for COVID-19 Online available: https://api.covid19india.org/documentation/csv/

38. List of states and union territories of India by population (Census 2011), https://en.wikipedia.org/wiki/List_of_states_and_union_territories_of_India_by_population

39. Population and decadal change by residence : 2011 (PERSONS) *(PDF). Office of the Registrar General & Census Commissioner, India. p. 2.* https://www.censusindia.gov.in/2011census/PCA/PCA_Highlights/pca_highlights_file/India/Chapter-1.pdf/

40. Python Software Foundation. Python Language Reference, version 2.7. Available: http://www.python.org/

41. Gupta, S. C., & Kapoor, V. K. (1994). Fundamental of Mathematical Statistics. Sultan Chand & Sons.

42. Uttar Pradesh orders 2-day weekend lockdown from today evening, https://www.hindustantimes.com/lucknow/uttar-pradesh-orders-2-day-weekend-lockdown-from-tomorrow-evening/story-85qzm3lFAOW1k8uoPLBLYI.html/

43. Covid-19: Cases soar to 33,053 in Maharashtra, nearly one-third of national total. https://www.hindustantimes.com/india-news/covid-19-state-tally-cases-soar-to-33-053-in-maharashtra-nearly-one-third-of-national-total/story-kPYzhFYmabDYxmwSxSwoDK.html/

44. Maharashtra extends Covid-19 lockdown till 30 June with some relaxations/ https://www.livemint.com/news/india/coronavirus-update-maharashtra-extends-covid-19-lockdown-till-30-june-with-some-relaxations-11590923986560.html/

45. Maharashtra extends lockdown till Aug 31 in view of rising Coronavirus cases. http://newsonair.com/News?title=Maharashtra-extends-lockdown-till-Aug-31-in-view-of-rising-Coronavirus-cases&id=396063/

46. Lockdown in Tamil Nadu extended till August 31. http://newsonair.com/Main-News-Details.aspx?id=396093/

47. West Bengal extends lockdown till July 31. https://www.thehindu.com/news/national/other-states/coronavirus-west-bengal-extends-lockdown-till-july-31/article31909219.ece/

48. West Bengal Lockdown: Mamata Extends Lockdown in Containment Zones Till August 31. https://www.india.com/news/india/west-bengal-lockdown-mamata-extends-lockdown-in-containment-zones-till-august-31-read-here-4098578/

49. COVID-19 pandemic in Bihar. https://en.wikipedia.org/wiki/COVID-19_pandemic_in_Bihar/

50. COVID-19 pandemic in Assam. https://en.wikipedia.org/wiki/COVID-19_pandemic_in_Assam/